\documentclass[technote]{IEEEtran}

\usepackage{graphicx}

\usepackage{amsmath}
\usepackage{amssymb}
\usepackage{amsfonts}

\usepackage{booktabs}
\usepackage{multirow}
\usepackage{subfig}
\usepackage{cite}
\usepackage{floatflt}
\usepackage{wrapfig}
\usepackage[linesnumbered, ruled,vlined]{algorithm2e}
\usepackage{moresize}
\usepackage{url}
\usepackage{xr}
\usepackage{tablefootnote}
\usepackage{bigfoot}
\usepackage[whole]{bxcjkjatype}

\usepackage{amsthm}
\theoremstyle{definition}
\newtheorem{definition}{Definition}

\usepackage[hidelinks,pdftitle={A Review of Evolutionary Multi-modal Multi-objective Optimization},pdfauthor={Ryoji Tanabe and Hisao Ishibuchi}]{hyperref}

\def\vector#1{\mbox{\boldmath $#1$}}

\ifCLASSINFOpdf
\else
\fi
\begin{document}
%


\title{A Review of Evolutionary Multi-modal Multi-objective Optimization}


%
%
%

\author{
Ryoji~Tanabe,~\IEEEmembership{Member,~IEEE,}and~Hisao~Ishibuchi,~\IEEEmembership{Fellow,~IEEE}
\thanks{R. Tanabe and H. Ishibuchi are with
Shenzhen Key Laboratory of Computational Intelligence, University Key Laboratory of Evolving Intelligent Systems of Guangdong Province, Department of Computer Science and Engineering, Southern University of Science and Technology, Shenzhen 518055, China. e-mail: (rt.ryoji.tanabe@gmail.com, hisao@sustc.edu.cn). (Corresponding author: Hisao Ishibuchi)}
}

\maketitle


\maketitle


\begin{abstract}


Multi-modal multi-objective optimization aims to find all Pareto optimal solutions including overlapping solutions in the objective space.
Multi-modal multi-objective optimization has been investigated in the evolutionary computation community since 2005.
However, it is difficult to survey existing studies in this field because they have been independently conducted and do not explicitly use the term ``multi-modal multi-objective optimization''.
To address this issue, this paper reviews existing studies of evolutionary multi-modal multi-objective optimization, including studies published under names that are different from ``multi-modal multi-objective optimization''.
Our review also clarifies open issues in this research area.

\end{abstract}


%
\IEEEpeerreviewmaketitle

\section{Introduction}
\label{sec:introduction}



A multi-objective evolutionary algorithm (MOEA) is an efficient optimizer for a multi-objective optimization problem (MOP) \cite{Deb01}.
MOEAs aim to find a non-dominated solution set that approximates the Pareto front in the objective space.
The set of non-dominated solutions found by an MOEA is usually used in an ``a posteriori'' decision-making process \cite{Miettinen98}.
A decision maker selects a final solution from the solution set according to her/his preference.







\begin{figure}[t]
\newcommand{\widthvar}{0.4}
  \begin{center} 
    \includegraphics[width=\widthvar\textwidth]{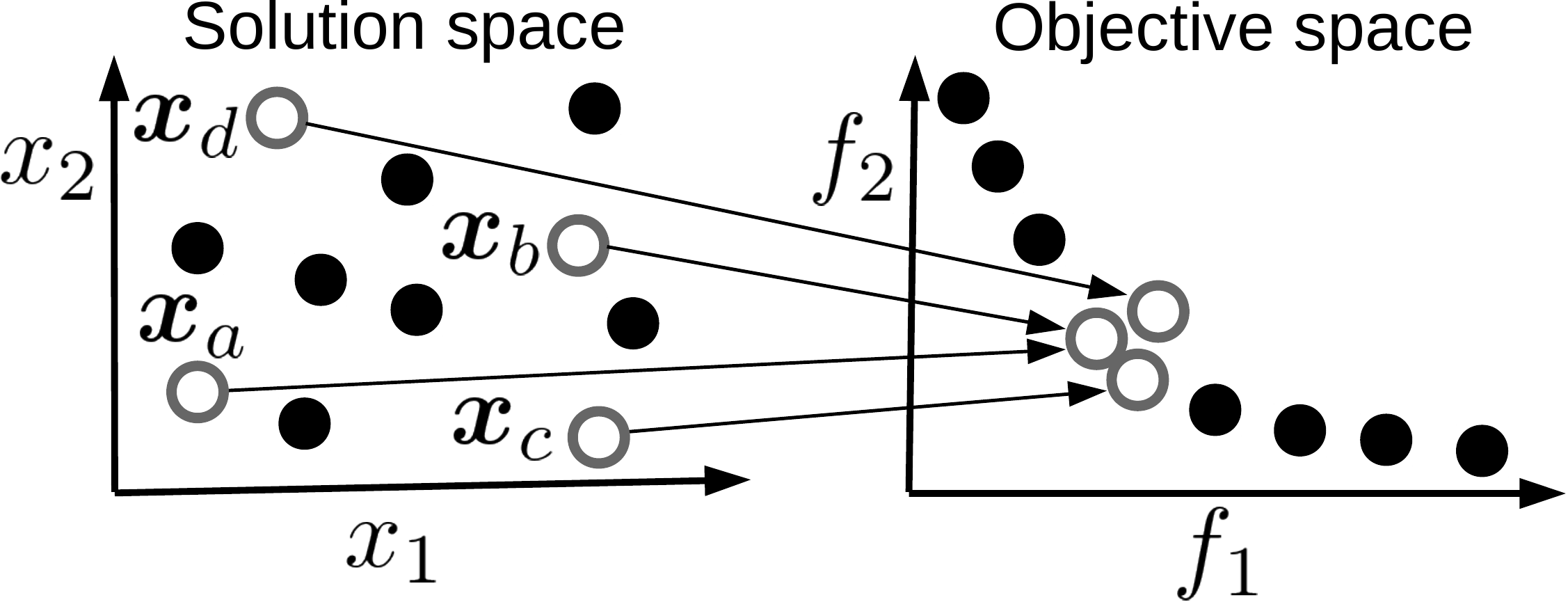}
\caption{
\small
Illustration of a situation where the four solutions are identical or close to each other in the objective space but are far from each other in the solution space (a minimization problem).
}
\label{fig:mmop_example}
  \end{center}
\end{figure}



Since the quality of a solution set is usually evaluated in the objective space, the distribution of solutions in the solution space has not received much attention in the evolutionary multi-objective optimization (EMO) community.
However, the decision maker may want to compare the final solution to other dissimilar solutions that have an equivalent quality or a slightly inferior quality \cite{SebagTTLB05,SchutzeVC11}.
Fig. \ref{fig:mmop_example} shows a simple example. 
In Fig. \ref{fig:mmop_example}, the four solutions $\vector{x}_a$, $\vector{x}_b$, $\vector{x}_c$, and $\vector{x}_d$ are far from each other in the solution space but close to each other in the objective space.
$\vector{x}_a$ and $\vector{x}_b$ have the same objective vector.
$\vector{x}_c$ and $\vector{x}_a$ are similar in the objective space.
$\vector{x}_d$ is dominated by these solutions. 
This kind of situation can be found in a number of real-world problems, including functional brain imaging problems \cite{SebagTTLB05}, diesel engine design problems \cite{HiroyasuNM05}, distillation plant layout problems \cite{PreussKBH08}, rocket engine design problems \cite{KudoYF11}, and game map generation problems \cite{TogeliusPY10}.



%

%

If multiple diverse solutions with similar objective vectors like $\vector{x}_a$, $\vector{x}_b$, $\vector{x}_c$, and $\vector{x}_d$ in Fig. \ref{fig:mmop_example} are obtained, the decision maker can select the final solution according to her/his preference in the solution space.
For example, if $\vector{x}_a$ in Fig. \ref{fig:mmop_example} becomes unavailable for some reason (e.g., material shortages, mechanical failures, traffic accidents, and law revisions), the decision maker can select a substitute from $\vector{x}_b$, $\vector{x}_c$, and $\vector{x}_d$.


A practical example is given in \cite{SchutzeVC11}, which deals with two-objective space mission design problems.
In \cite{SchutzeVC11}, Sch{\"{u}}tze et al. considered two dissimilar solutions $\vector{x}_1 = (782, 1288, 1788)^{\rm T}$ and $\vector{x}_2 = (1222, 1642, 2224)^{\rm T}$ for a minimization problem, whose objective vectors are $\vector{f}(\vector{x}_1) = (0.462, 1001.7)^{\rm T}$ and $\vector{f}(\vector{x}_2) = (0.463, 1005.3)^{\rm T}$, respectively.
Although $\vector{x}_1$ dominates $\vector{x}_2$, the difference between $\vector{f}(\vector{x}_1)$ and $\vector{f}(\vector{x}_2)$ is small enough.
The first design variable is the departure time from the Earth (in days).
Thus, the departure times of $\vector{x}_2$ and $\vector{x}_1$ differ by $440$ days ($= 1222 - 782$).
If the decision maker accepts $\vector{x}_2$ with a slightly inferior quality in addition to $\vector{x}_1$, the two launch plans can be considered.
If $\vector{x}_1$ is not realizable for some reason, $\vector{x}_2$ can be the final solution instead of $\vector{x}_1$.
As explained here,
multiple solutions with almost equivalent quality support a reliable decision-making process.
If these solutions have a large diversity in the solution space, they can provide insightful information for engineering design  \cite{SebagTTLB05,HiroyasuNM05}.

\begin{figure}[t]
\newcommand{\widthvar}{0.21}
\centering
\subfloat[$\vector{F}$]{\includegraphics[width=0.194\textwidth]{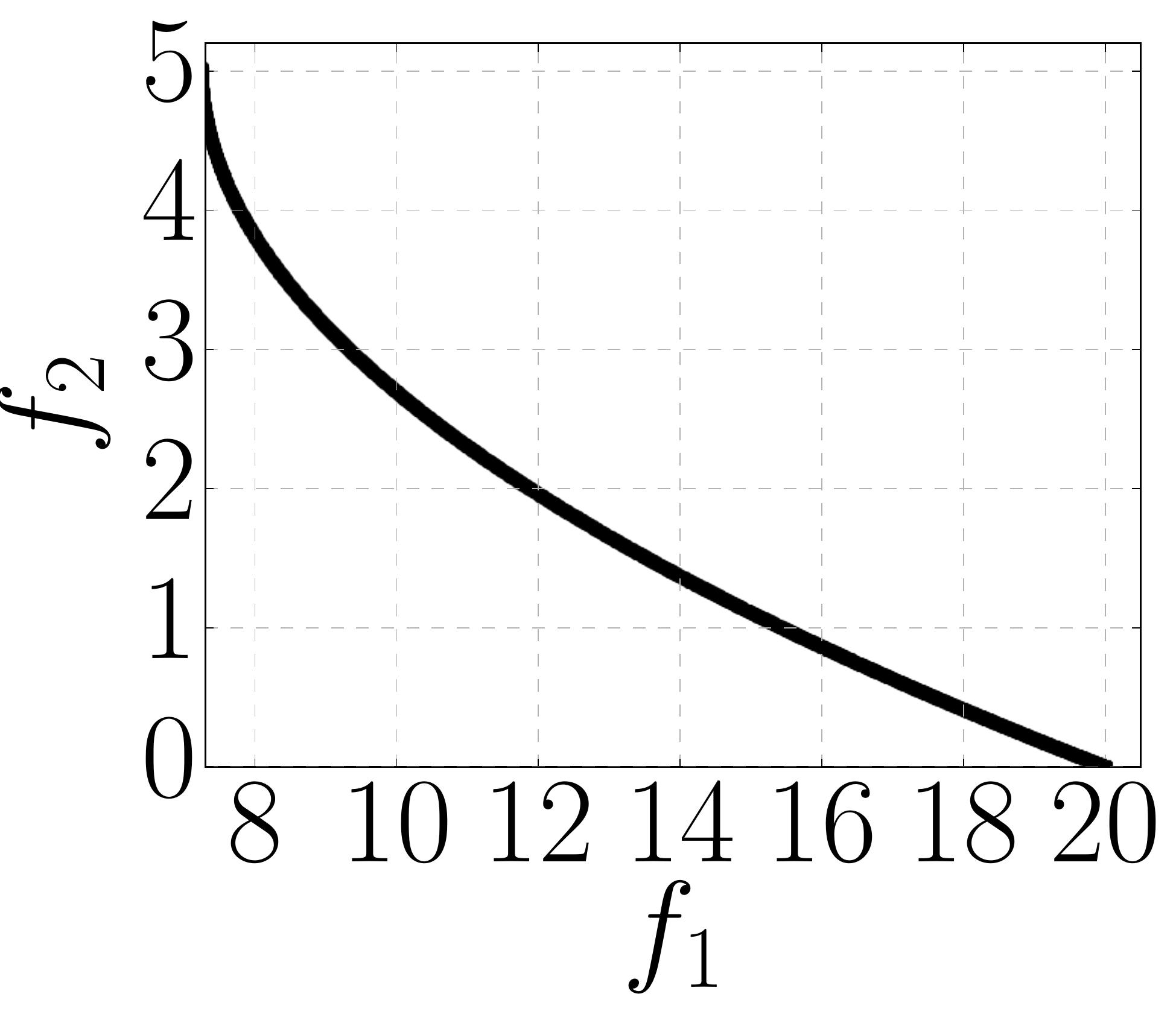}}
\subfloat[$\vector{O}$]{\includegraphics[width=\widthvar\textwidth]{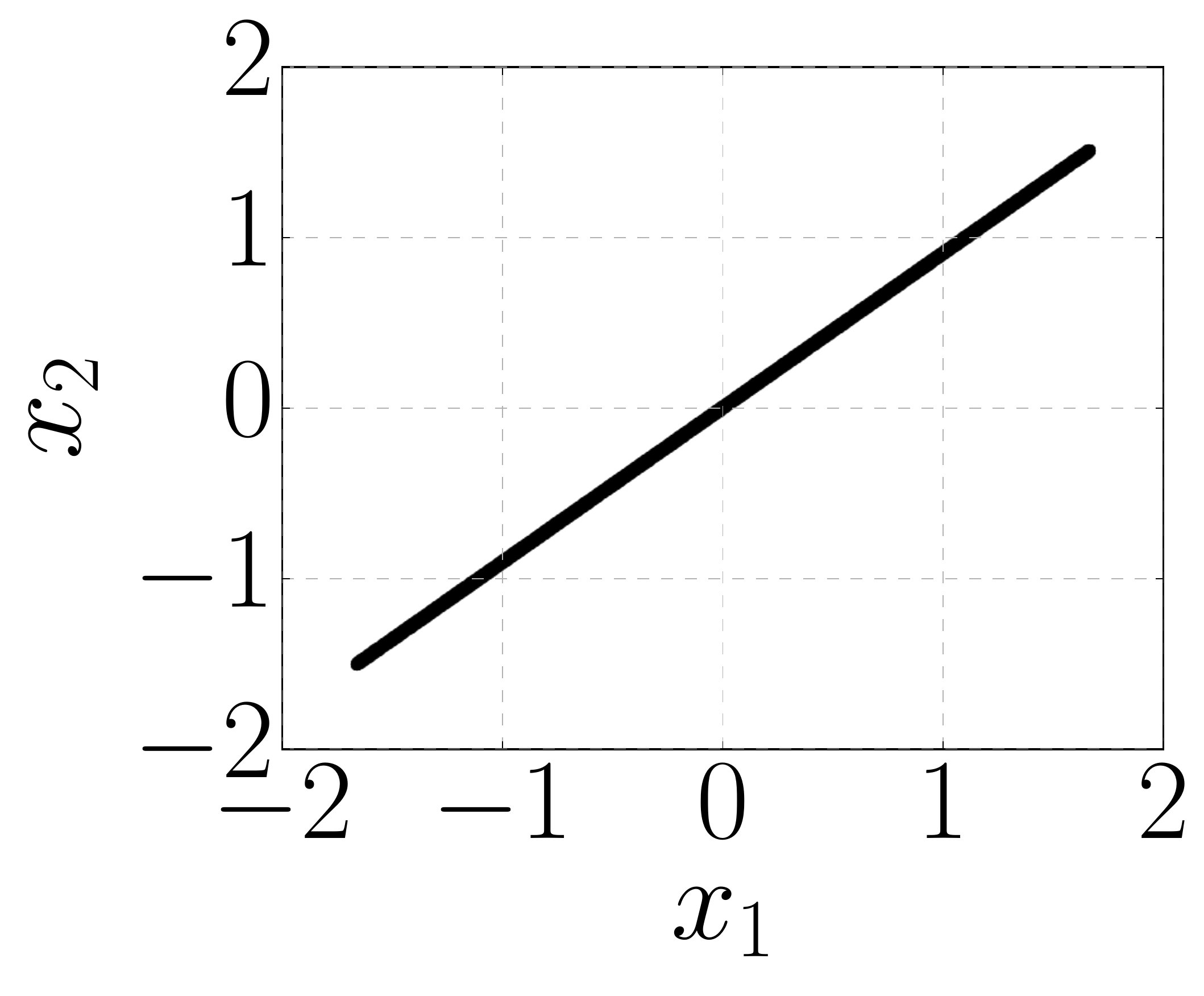}}
\\[-3mm]
\subfloat[$\vector{O}_1$]{\includegraphics[width=\widthvar\textwidth]{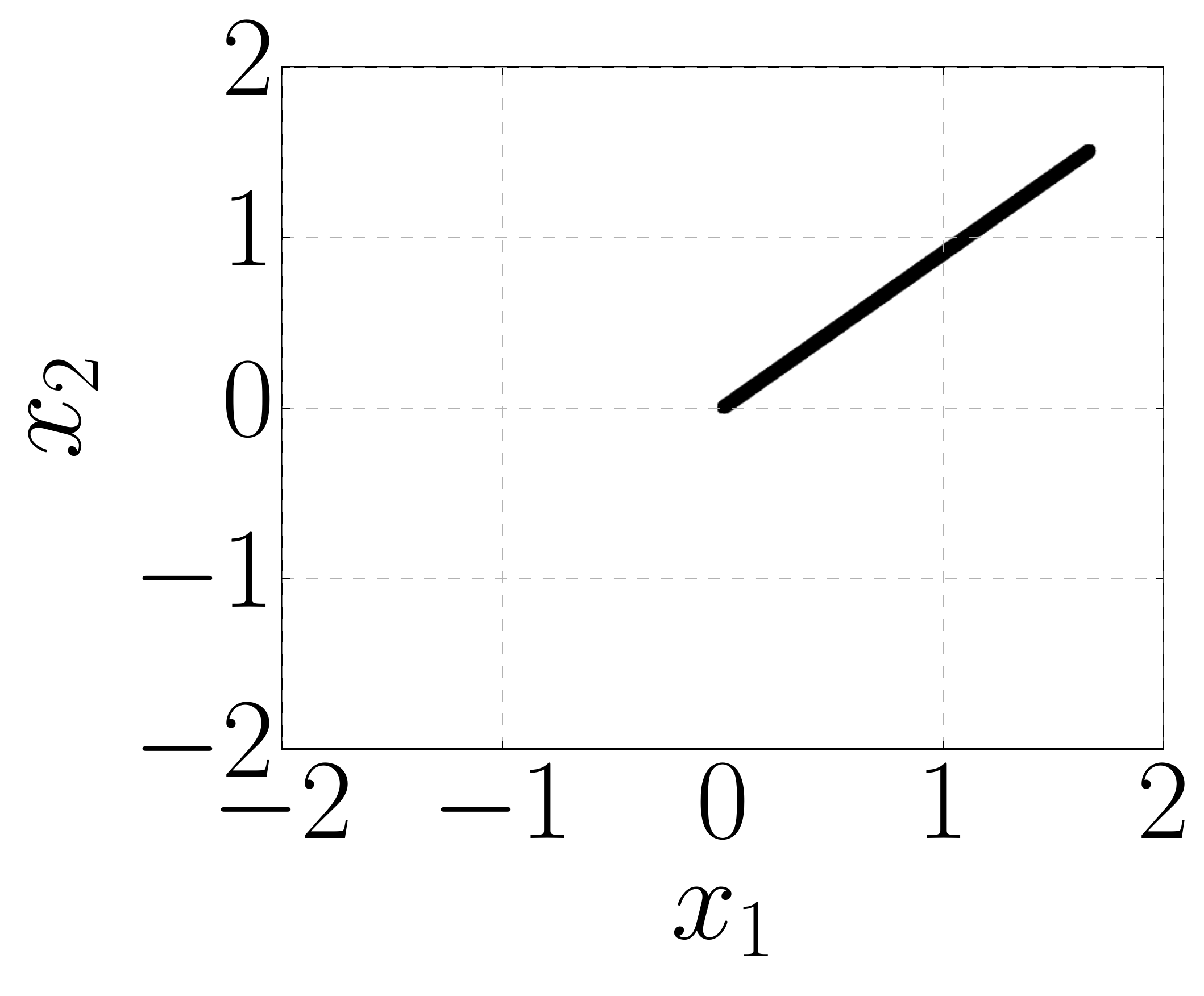}}
\subfloat[$\vector{O}_2$]{\includegraphics[width=\widthvar\textwidth]{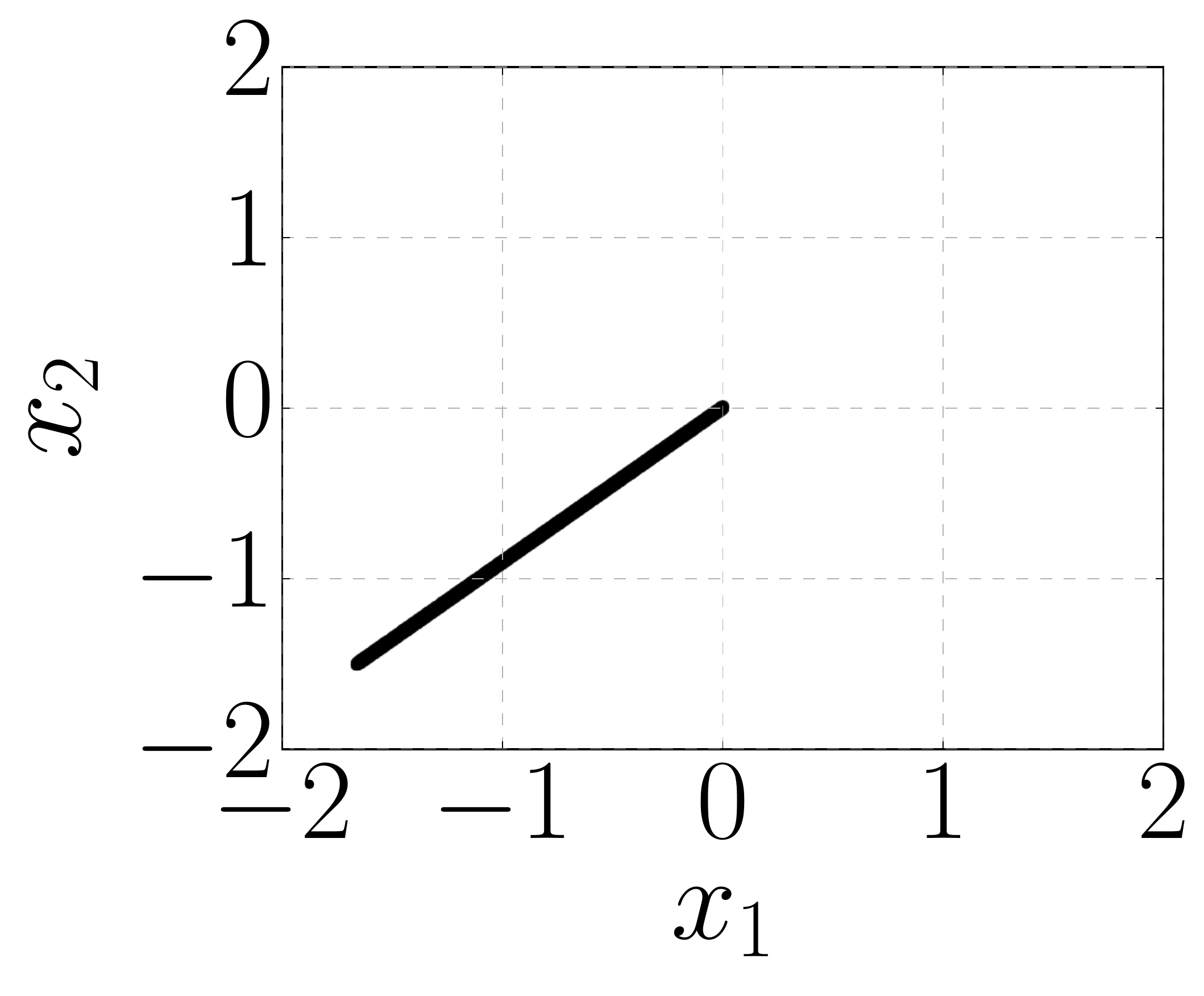}}
\caption{
\small
%
(a) The Pareto front $\vector{F}$ and (b) the Pareto optimal solution set $\vector{O}$ of Two-On-One \cite{PreussNR06}.
Figs. (c) and (d) show the two Pareto optimal solution subsets $\vector{O}_1$ and $\vector{O}_2$, respectively.
}
\label{fig:two-on-one}
\end{figure}

A multi-modal multi-objective optimization problem (MMOP) involves finding all solutions that are equivalent to Pareto optimal solutions \cite{SebagTTLB05,DebT08,LiEDE17}.
Below, we explain the difference between MOPs and MMOPs using the two-objective and two-variable Two-On-One problem \cite{PreussNR06}.
Figs. \ref{fig:two-on-one} (a) and (b) show the Pareto front $\vector{F}$ and the Pareto optimal solution set $\vector{O}$ of Two-On-One, respectively.
Two-On-One has two equivalent Pareto optimal solution subsets $\vector{O}_1$ and $\vector{O}_2$ that are symmetrical with respect to the origin, where $\vector{O} = \vector{O}_1 \cup \vector{O}_2$.
Figs. \ref{fig:two-on-one} (c) and (d) show $\vector{O}_1$ and $\vector{O}_2$, respectively.
In Two-On-One, the three solution sets $\vector{O}$, $\vector{O}_1$, and $\vector{O}_2$ (Figs. \ref{fig:two-on-one} (b), (c) and (d)) are mapped to $\vector{F}$ (Fig. \ref{fig:two-on-one} (a)) by the objective functions.
On the one hand, the goal of MOPs is generally to find a solution set that approximates the Pareto front $\vector{F}$ in the objective space.
Since $\vector{O}_1$ and $\vector{O}_2$ are mapped to the same $\vector{F}$ in the objective space, it is sufficient for MOPs to find either $\vector{O}_1$ or $\vector{O}_2$.
On the other hand, the goal of MMOPs is to find the entire equivalent Pareto optimal solution set $\vector{O} = \vector{O}_1 \cup \vector{O}_2$ in the solution space.
In contrast to MOPs, it is necessary to find both $\vector{O}_1$ and $\vector{O}_2$ in MMOPs.
Since most MOEAs (e.g., NSGA-II \cite{DebAPM02} and SPEA2 \cite{ZitzlerLT01}) do not have mechanisms to maintain the solution space diversity, it is expected that they do not work well for MMOPs.
Thus, multi-modal multi-objective evolutionary algorithms (MMEAs) that handle the solution space diversity are necessary for MMOPs.

This paper presents a review of evolutionary multi-modal multi-objective optimization.
This topic is not new and has been studied for more than ten years.
Early studies include \cite{DebT05,HiroyasuNM05,KimHMW04,SebagTTLB05,PreussNR06,RudolphNP07}.
Unfortunately, most existing studies were independently conducted and did not use the term ``MMOPs'' (i.e., they are not tagged).
For this reason, it is difficult to survey existing studies of MMOPs despite their significant contributions.
In this paper, we review related studies of MMOPs including those published under names that were different from ``multi-modal multi-objective optimization''.
We also clarify open issues in this field.
Multi-modal single-objective optimization problems (MSOPs) have been well studied in the evolutionary computation community \cite{LiEDE17}.
Thus, useful clues to address some issues in studies of MMOPs may be found in studies of MSOPs.
We discuss what can be learned from the existing studies of MSOPs.









This paper is organized as follows.
Section \ref{sec:definition} gives definitions of MMOPs.
Section \ref{sec:emmo} describes MMEAs.
Section \ref{sec:test_problems} presents test problems for multi-modal multi-objective optimization.
Section \ref{sec:metrics} explains performance indicators for benchmarking MMEAs.
Section \ref{sec:conclusion} concludes this paper. 



\newtheorem{defi}{Definition}

\section{Definitions of MMOPs}
\label{sec:definition}





\subsubsection{Definition of MOPs}

A continuous MOP involves finding a solution $\vector{x} \in \mathbb{S} \subseteq \mathbb{R}^D$ that minimizes a given objective function vector $\vector{f}: \mathbb{S} \rightarrow \mathbb{R}^M$.
Here, $\mathbb{S}$ is the $D$-dimensional solution space, and $\mathbb{R}^M$ is the $M$-dimensional objective space.

A solution $\vector{x}_1$ is said to dominate $\vector{x}_2$ iff $f_i (\vector{x}_1) \leq f_i (\vector{x}_2)$ for all $i \in \{1, ..., M\}$ and $f_i (\vector{x}_1) < f_i (\vector{x}_2)$ for at least one index $i$.
If $ \vector{x}^*$ is not dominated by any other solutions, it is called a Pareto optimal solution.
The set of all $\vector{x}^*$ is the Pareto optimal solution set, and the set of all $\vector{f}(\vector{x}^*)$ is the Pareto front.
The goal of MOPs is generally to find a non-dominated solution set that approximates the Pareto front in the objective space.


\subsubsection{Definitions of MMOPs}
\label{sec:def_mmops}



The term ``MMOP'' was first coined in \cite{DebT05,SebagTTLB05} in 2005.
However, ``MMOP'' was not used in most studies from 2007 to 2012.
Terms that represent MMOPs were not explicitly defined in those studies.
For example, MMOPs were referred to as problems of obtaining a diverse solution set in the solution space in \cite{RudolphP09}.
It seems that ``multi-modal multi-objective optimization'' has been used again as of 2016.
Apart from these instances, MMOPs were denoted as ``Multi-objective multi-global optimization'' and ``Multi-modal multi-objective wicked problems'' in \cite{CoelhoZ06} and \cite{ZechmanGS13}, respectively.


Although MMOPs have been addressed for more than ten years, the definition of an MMOP is still controversial.
In this paper, we define an MMOP using a relaxed equivalency introduced by Rudolph and Preuss \cite{RudolphP09} as follows:

\begin{definition}
An MMOP involves finding all solutions that are equivalent to Pareto optimal solutions.
\end{definition} 

\begin{definition}
Two different solutions $\vector{x}_1$ and $\vector{x}_2$ are said to be equivalent iff $\|\vector{f}(\vector{x}_1) - \vector{f}(\vector{x}_2) \| \leq \delta$.
\end{definition}

\noindent where $\| \vector{a} \|$ is an arbitrary norm of $\vector{a}$, and $\delta$ is a non-negative threshold value given by the decision maker.
If $\delta = 0$, the MMOP should find all equivalent Pareto optimal solutions.
If $\delta > 0$, the MMOP should find all equivalent Pareto optimal solutions and dominated solutions with acceptable quality.
The main advantage of our definition of an MMOP is that the decision maker can adjust the goal of the MMOP by changing the $\delta$ value.
Most existing studies (e.g., \cite{DebT08,ShirPNE09,YueQL17}) assume MMOPs with $\delta=0$.
MMOPs with $\delta>0$ were discussed in \cite{SebagTTLB05,SchutzeVC11,UlrichBT10,ZechmanGS13}.
For example, $\vector{x}_a$, $\vector{x}_b$, and $\vector{x}_c$ in Fig. \ref{fig:mmop_example} should be found for MMOPs with $\delta=0$.
In addition, the non-Pareto optimal solution $\vector{x}_d$ should be found for MMOPs with $\delta > 0$ if $\| \vector{f}(\vector{x}_d) - \vector{f}(\vector{x}_a)\| \leq \delta$.

Although there is room for discussion, MMOPs with $\delta>0$ may be more practical in real-world applications.
This is because the set of solutions of an MMOP with $\delta>0$ can provide more options for the decision maker than that of  an MMOP with $\delta=0$.
While it is usually assumed in the EMO community that the final solution is selected from non-dominated solutions, the decision maker may also be interested in some dominated solutions in practice \cite{SebagTTLB05,SchutzeVC11}.
Below, we use the term ``MMOP'' regardless of the $\delta$ value for simplicity.

\section{MMEAs}
\label{sec:emmo}

This section describes 12 dominance-based MMEAs, 3 decomposition-based MMEAs, 2 set-based MMEAs, and a post-processing approach.
MMEAs need the following three abilities: (1) the ability to find solutions with high quality, (2)  the  ability to find diverse solutions in the objective space, and (3) the ability to find diverse solutions in the solution space.
MOEAs need the abilities (1) and (2) to find a solution set that approximates the Pareto front in the objective space.
Multi-modal single-objective optimizers need the abilities (1) and (3) to find a set of global optimal solutions.
In contrast, MMEAs need all abilities (1)--(3).
Here, we mainly describe mechanisms of each type of MMEA to handle (1)--(3).


\subsubsection{Pareto dominance-based MMEAs}

The most representative MMEA is Omni-optimizer \cite{DebT05,DebT08}, which is an NSGA-II-based generic optimizer applicable to various types of problems.
The differences between Omni-optimizer and NSGA-II are fourfold: the Latin hypercube sampling-based population initialization, the so-called restricted mating selection, the $\epsilon$-dominance-based non-dominated sorting, and the alternative crowding distance.
In the restricted mating selection, an individual $\vector{x}_a$ is randomly selected from the population.
Then, $\vector{x}_a$ and its nearest neighbor $\vector{x}_b$ in the solution space are compared based on their non-domination levels and crowding distance values.
The winner among $\vector{x}_a$ and $\vector{x}_b$ is selected as a parent.

The crowding distance measure in Omni-optimizer takes into account both the objective and solution spaces.
For the $i$-th individual $\vector{x}_i$ in each non-dominated front $\vector{R}$, the crowding distance in the objective space $c^{\rm obj}_{i}$ is calculated in a similar manner to NSGA-II.
In contrast, the crowding distance value of $\vector{x}_i$ in the solution space $c^{\rm sol}_{i}$ is calculated in a different manner.
First, for each $j \in \{1, ..., D\}$, a ``variable-wise'' crowding distance value of $\vector{x}_i$ in the $j$-th decision variable $c^{\rm sol}_{i, j}$ is calculated as follows:
\begin{align}
\label{eqn:variablewise_cd}
  c^{\rm sol}_{i, j} = \begin{cases}
    2 \left(\frac{x_{i+1, j} - x_{i, j}}{x_j^{\rm max} - x_j^{\rm min}}\right) & \:  {\rm if} \: x_{i,j} = x_j^{\rm min}\\
    2\left(\frac{x_{i, j} - x_{i-1, j}}{x_j^{\rm max} - x_j^{\rm min}}\right) & \:  {\rm else \: if} \: x_{i,j} = x_j^{\rm max}\\
   \frac{x_{i+1, j} - x_{i-1, j}}{x_j^{\rm max} - x_j^{\rm min}} & \:  {\rm otherwise}
     \end{cases},  
\end{align}  
where we assume that all individuals in $\vector{R}$ are sorted based on their $j$-th decision variable values in descending order.
In \eqref{eqn:variablewise_cd}, $x_j^{\rm min} = \min_{\vector{x} \in \vector{R}} \{x_j\}$ and $x_j^{\rm max} = \max_{\vector{x} \in \vector{R}} \{x_j\}$.
Unlike the crowding distance in the objective space, an infinitely large value is not given to a boundary individual.

Then, an ``individual-wise'' crowding distance value  $c^{\rm sol}_{i}$ is calculated as follows: $c^{\rm sol}_{i} = (\sum^{D}_{j=1} c^{\rm sol}_{i, j})/D$.
The average value $c^{\rm sol}_{\rm avg}$ of all individual-wise crowding distance values is also calculated as follows: $c^{\rm sol}_{\rm avg} = (\sum^{|\vector{R}|}_{i=1} c^{\rm sol}_{i}) / |\vector{R}|$.
Finally, the crowding distance value $c_{i}$ of $\vector{x}_i$ is obtained as follows:
\begin{align}
\label{eqn:omni-optimizer_crowding}
  c_{i} = \begin{cases}
    \max \{c^{\rm obj}_{i}, c^{\rm sol}_{i}\} & \:  {\rm if} \: c^{\rm obj}_{i} > c^{\rm obj}_{\rm avg} \: {\rm or} \: c^{\rm sol}_{i} > c^{\rm sol}_{\rm avg}\\
    \min \{c^{\rm obj}_{i}, c^{\rm sol}_{i}\} & \:  {\rm otherwise}
     \end{cases},  
\end{align}  
where $c^{\rm obj}_{\rm avg}$ is the average value of all crowding distance values in the objective space.
As shown in \eqref{eqn:omni-optimizer_crowding}, $c_{i}$ in Omni-optimizer is the combination of $c^{\rm obj}_{i}$ and $c^{\rm sol}_{i}$.
Due to its alternative crowding distance, the results presented in \cite{DebT08} showed that Omni-optimizer finds more diverse solutions than NSGA-II.



In addition to Omni-optimizer, two extensions of NSGA-II for MMOPs have been proposed.
DNEA \cite{LiuINMS18} is similar to Omni-optimizer but uses two sharing functions in the objective and solution spaces. 
DNEA requires fine-tuning of two sharing niche parameters for the objective and solution spaces.
%
The secondary criterion of DN-NSGA-II \cite{LiangYQ16} is based on the crowding distance only in the solution space.
DN-NSGA-II uses a solution distance-based mating selection.



The following are other dominance-based MMEAs. 
An MMEA proposed in \cite{KramerD10} utilizes DBSCAN \cite{EsterKSX96} and the rake selection \cite{KramerK09}.
DBSCAN, which is a clustering method, is used for grouping individuals based on the distribution of individuals in the solution space.
The rake selection, which is a reference vector-based selection method similar to NSGA-III \cite{DebJ14}, is applied to individuals belonging to each niche for the environmental selection.
SPEA2$+$ \cite{HiroyasuNM05,KimHMW04} uses two archives $\vector{A}_{\rm obj}$ and $\vector{A}_{\rm sol}$ to maintain diverse non-dominated individuals in the objective and solution spaces, respectively.
While the environmental selection in $\vector{A}_{\rm obj}$ is based on the density of individuals in the objective space similar to SPEA2 \cite{ZitzlerLT01}, that in $\vector{A}_{\rm sol}$ is based on the density of individuals in the solution space.
For the mating selection in SPEA2$+$, neighborhood individuals in the objective space are selected only from $\vector{A}_{\rm obj}$.

$P_{Q, \epsilon}$-MOEA \cite{SchutzeVC11}, 4D-Miner \cite{SebagTTLB05,KrmicekS06}, and MNCA \cite{ZechmanGS13} are capable of handling dominated solutions for MMOPs with $\delta>0$.
$P_{Q, \epsilon}$-MOEA uses the $\epsilon$-dominance relation \cite{LaumannsTDZ02} so that an unbounded archive can maintain individuals with acceptable quality according to the decision maker.
Unlike other MMEAs, $P_{Q, \epsilon}$-MOEA does not have an explicit mechanism to maintain the solution space diversity.
4D-Miner was specially designed for functional brain imaging problems \cite{SebagTTLB05}.
The population is initialized by a problem-specific method. 
4D-Miner maintains dissimilar individuals in an external archive, whose size is ten times larger than the population size.
The environmental selection in 4D-Miner is based on a problem-specific metric.
Similar to DIOP \cite{UlrichBT10} (explained later), MNCA simultaneously evolves multiple subpopulations $\vector{P}_1, ..., \vector{P}_S$, where $S$ is the number of subpopulations.
In MNCA, the primary subpopulation $\vector{P}_1$ aims to find an approximation of the Pareto front that provides a target front for other subpopulations $\vector{P}_2, ..., \vector{P}_S$.
While the update of $\vector{P}_1$ is based on the same selection mechanism as in NSGA-II,
the update of $\vector{P}_2, ..., \vector{P}_S$ is performed with a complicated method that takes into account both the objective and solution spaces.

Although the above-mentioned MMEAs use genetic variation operators (e.g., the SBX crossover and the polynomial mutation \cite{DebAPM02}), the following MMEAs are based on other approaches.
Niching-CMA \cite{ShirPNE09} is an extension of CMA-ES \cite{HansenO01} for MMOPs by introducing a niching mechanism.
The number of niches and the niche radius are adaptively adjusted in Niching-CMA.
An aggregate distance metric in the objective and solution spaces is used to group individuals into multiple niches.
For each niche, individuals with better non-domination levels survive to the next iteration.
MO\_Ring\_PSO\_SCD \cite{YueQL17}, a PSO algorithm for MMOPs, uses a diversity measure similar to Omni-optimizer.
However, MO\_Ring\_PSO\_SCD handles the boundary individuals in the objective space in an alternative manner.
In addition, an index-based ring topology is used to create niches.


Two extensions of artificial immune systems \cite{DasguptaYN11} have been proposed for MMOPs: omni-aiNet \cite{CoelhoZ06} and cob-aiNet \cite{CoelhoZ11}.
These two methods use a modified version of the polynomial mutation \cite{DebAPM02}.
The primary and secondary criteria of omni-aiNet are based on $\epsilon$-nondomination levels \cite{LaumannsTDZ02} and a grid operation, respectively.
In addition, omni-aiNet uses suppression and insertion operations. 
While the suppression operation deletes an inferior individual, the insertion operation adds new individuals to the population.
The population size is not constant due to these two operations.
The primary and secondary criteria of cob-aiNet are based on the fitness assignment method in SPEA2 \cite{ZitzlerLT01} and a diversity measure with a sharing function in the solution space, respectively.
The maximum population size is introduced in cob-aiNet.





\subsubsection{Decomposition-based MMEAs}



%
A three-phase multi-start method is proposed in \cite{RudolphNP07}.
First, $(1, \lambda)$-ES is carried out on each $M$ objective functions $K$ times to obtain $M \times K$ best-so-far solutions.
Then, an unsupervised clustering method is applied to the $M \times K$ solutions to detect the number of equivalent Pareto optimal solution subsets $s$. 
Finally,  $s$ runs of $(1, \lambda)$-ES are performed on each $N$ single-objective subproblem decomposed by the Tchebycheff function.
The initial individual of each run is determined in a chained manner.
The best solution found in the $j$-th subproblem becomes an initial individual of $(1, \lambda)$-ES for the $j+1$-th subproblem ($j \in \{1, ..., N-1\}$).
It is expected that $s$ equivalent solutions are found for each $N$ decomposed subproblems.



Two variants of MOEA/D \cite{ZhangL07} for MMOPs are proposed in \cite{HuI18,TanabeI18}. 
MOEA/D decomposes an $M$-objective problem into $N$ single-objective subproblems using a set of weight vectors, assigning a single individual to each subproblem.
Then, MOEA/D simultaneously evolves the $N$ individuals.
Unlike MOEA/D, the following two methods assign one or more individuals to each subproblem to handle the equivalency.

The MOEA/D algorithm presented in \cite{HuI18} assigns $K$ individuals to each subproblem.
The selection is conducted based on a fitness value combining the PBI function value \cite{ZhangL07} and two distance values in the solution space.
$K$ dissimilar individuals are likely to be assigned to each subproblem.
The main drawback of the above methods \cite{HuI18,RudolphNP07} is the difficulty in setting a proper value for $K$, because it is problem dependent.
MOEA/D-AD \cite{TanabeI18} does not need such a parameter but requires a relative neighborhood size $L$. 
For each iteration, a child $\vector{u}$ is assigned to the $j$-th subproblem whose weight vector is closest to $\vector{f}(\vector{u})$, with respect to the perpendicular distance.
Let $\vector{X}$ be a set of individuals already assigned to the $j$th-subproblem.
If $\vector{x}$ in $\vector{X}$ is within the $L$ nearest individuals from the child $\vector{u}$ in the solution space, $\vector{x}$ and $\vector{u}$ are compared based on their scalarizing function values $g(\vector{x})$ and $g(\vector{u})$.
If $g(\vector{u}) \leq g(\vector{x})$, $\vector{x}$ is deleted from the population and $\vector{u}$ enters the population.
$\vector{u}$ also enters the population when no $\vector{x}$ in $\vector{X}$ is in the $L$ neighborhood of $\vector{u}$ in the solution space.



\subsubsection{Set-based MMEAs}

DIOP \cite{UlrichBT10} is a set-based MMEA that can maintain dominated solutions in the population.
In the set-based optimization framework \cite{ZitzlerTB10}, a single solution in the upper level represents a set of solutions in the lower level (i.e., a problem).
%
DIOP simultaneously evolves an archive $\vector{A}$ and a target population $\vector{T}$.
While $\vector{A}$ approximates only the Pareto front and is not shown to the decision maker, $\vector{T}$ obtains diverse solutions with acceptable quality by maximizing the following $G$ indicator: $G(\vector{T}) = w_{\rm obj} D_{\rm obj} (\vector{T}) + w_{\rm sol} D_{\rm sol} (\vector{T})$.
Here, $w_{\rm obj} + w_{\rm sol} = 1$.
$D_{\rm obj}$ is a performance indicator in the objective space, and $D_{\rm sol}$ is a diversity measure in the solution space.
In \cite{UlrichBT10}, $D_{\rm obj}$ and $D_{\rm sol}$ were specified by the hypervolume indicator \cite{ZitzlerTLFF03} and the Solow-Polasky diversity measure \cite{SolowP94}, respectively.
Meta-individuals in $\vector{T}$ that are $\epsilon$-dominated by any meta-individuals in $\vector{A}$ are excluded for the calculation of the $G$ metric.
At the end of the search, $\vector{T}$ is likely to contain meta-individuals (i.e., solution sets of a problem) $\epsilon$-nondominated by meta-individuals in $\vector{A}$.


Another set-based MMEA is presented in \cite{IshibuchiYAN12}.
Unlike DIOP, the proposed method evolves only a single population.
Whereas DIOP maximizes the weighted sum of values of $D_{\rm obj}$ and $D_{\rm sol}$, the proposed method treats $D_{\rm obj}$ and $D_{\rm sol}$ as meta two-objective functions.
NSGA-II is used to simultaneously maximize $D_{\rm obj}$ and $D_{\rm sol}$ in \cite{IshibuchiYAN12}.

\subsubsection{A post-processing approach}

As pointed out in \cite{RudolphP09}, it is not always necessary to locate all Pareto optimal solutions.
Suppose that a set of non-dominated solutions $\vector{A}$ has already been obtained by an MOEA (e.g., NSGA-II) but not an MMEA (e.g., Omni-optimizer).
After the decision maker has selected the final solution $\vector{x}_{\rm final}$ from $\vector{A}$ according to her/his preference in the objective space, it is sufficient to search solutions whose objective vectors are equivalent to $\vector{f}(\vector{x}_{\rm final})$.

A post-processing approach is proposed in \cite{RudolphP09} to handle this problem.
First, the proposed approach formulates a meta constrained two-objective minimization problem where $f^{\rm meta}_1 = \|\vector{f}(\vector{x}) - \vector{f}(\vector{x}_{\rm final})\|^2$, $f^{\rm meta}_2 = -\|\vector{x} - \vector{x}_{\rm final}\|^2$, and $g^{\rm meta}(\vector{x}) = f^{\rm meta}_1(\vector{x}) - \theta < 0$.
The meta objective functions $f^{\rm meta}_1$ and $f^{\rm meta}_2$ represent the distance between $\vector{x}$ and $\vector{x}_{\rm final}$ in the objective and solution spaces.
Thus, smaller $f^{\rm meta}_1 (\vector{x})$ and $f^{\rm meta}_2 (\vector{x})$ indicate that $\vector{x}$ is similar to $\vector{x}_{\rm final}$ in the objective space and far from $\vector{x}_{\rm final}$ in the solution space, respectively.
The constraint $g^{\rm meta}$ with $\theta > 0$ prevents $f^{\rm meta}_2 (\vector{x})$ from becoming an infinitely small value in unbounded problems.
NSGA-II is used as a meta-optimizer in \cite{RudolphP09}.

\begin{table}[t]
\renewcommand{\arraystretch}{0.8}
\begin{center}
  \caption{\small Properties of 18 MMEAs. $\mu$ and $n_{\rm max}$ denote the population size and the maximum number of evaluations used in each paper, respectively. ``$\delta>0$'' indicates whether each method can handle MMOPs with $\delta>0$. ``U'' means whether each method has an unbounded population/archive. Initial $\mu$ values are reported for omni-aiNet, cob-aiNet, $P_{Q, \epsilon}$-MOEA, and MOEA/D-AD. $\mu$ and $n_{\rm max}$ used in the post-processing step are shown for a method in \cite{RudolphP09}.}
{\scriptsize
  \label{tab:emmoas}
\scalebox{1}[0.98]{ 
\begin{tabular}{llcccccccc}
\midrule
& MMEAs & Year & $\mu$ & $n_{\rm max}$ & $\delta>0$ & U \\
\toprule
& SPEA2$+$ \cite{HiroyasuNM05,KimHMW04} & 2004 &  $100$ & $ 50\,000$ &  &  \\\cmidrule{2-7}
& Omni-optimizer \cite{DebT05,DebT08} & 2005 & $1\,000$ & $500\,000$ &  &  \\\cmidrule{2-7}
& 4D-Miner \cite{SebagTTLB05,KrmicekS06}  & 2005 & $200$ &  $8\,000$  & $\checkmark$ &  \\\cmidrule{2-7}
& omni-aiNet \cite{CoelhoZ06} & 2006 & $400$ & $40\,000$ &  &  $\checkmark$ \\\cmidrule{2-7}
& Niching-CMA \cite{ShirPNE09} & 2009 & $50$ & $50\,000$ &  &  \\\cmidrule{2-7}
\multirow{2}{*}{\rotatebox{90}{Dominance}}  & A method in \cite{KramerD10} & 2010 &\multicolumn{2}{c}{Not clearly reported}   &  &  \\\cmidrule{2-7}
& $P_{Q, \epsilon}$-MOEA \cite{SchutzeVC11} & 2011& $200$ & $5\,000$ & $\checkmark$ & $\checkmark$ \\\cmidrule{2-7}
& cob-aiNet \cite{CoelhoZ11} & 2011 & $100$ & $40\,000$ &  &  \\\cmidrule{2-7}
& MNCA \cite{ZechmanGS13} & 2013 &  $100$ &  $100\,000$ & $\checkmark$ &  \\\cmidrule{2-7}
& DN-NSGA-II \cite{LiangYQ16} & 2016 &  $800$  & $80\,000$ &  &  \\\cmidrule{2-7}
& MO\_Ring\_PSO\_SCD \cite{YueQL17} & 2017 & $800$  & $80\,000$ &  &  \\\cmidrule{2-7}
& DNEA \cite{LiuINMS18} & 2018 & $210$ &$63\,000$  &  &  \\\midrule
\multirow{2}{*}{\rotatebox{90}{Decomp.}} & A  method in \cite{RudolphNP07} & 2007 & $10$ & $20\,000$ &  &  \\\cmidrule{2-7}
& A method in \cite{HuI18}& 2018 & $1\,120$ & $89\,600$ &  &  \\\cmidrule{2-7}
& MOEA/D-AD \cite{TanabeI18} & 2018 & $100$ &$30\,000$ & & $\checkmark$ \\\midrule
\multirow{1}{*}{\rotatebox{90}{Set\,\,\,}}  & DIOP \cite{UlrichBT10} & 2010 & $50$ & $100\,000$ & $\checkmark$  &  \\\cmidrule{2-7}
& A method in \cite{IshibuchiYAN12} & 2012 & $200$ & $400\,000$ &  &  \\\midrule
\multirow{1}{*}{\rotatebox{90}{P.}} & A  method in \cite{RudolphP09} & 2009 & $20$  & $2\,000$ &  &  \\\midrule
\end{tabular}
}
}
 \end{center}
\end{table}

\subsubsection{Open issues}



Table \ref{tab:emmoas} summarizes the properties of the 18 MMEAs reviewed in this section.

While some MMEAs require an extra parameter (e.g., $L$ in MOEA/D-AD), Omni-optimizer does not require such a parameter.
This parameter-less property is an advantage of Omni-optimizer.
However, Omni-optimizer is a Pareto dominance-based MMEA.
Since dominance-based MOEAs perform poorly on most MOPs with more than three objectives \cite{DebJ14}, Omni-optimizer is unlikely to handle many objectives.

%
%
In addition to MMEAs, some MOEAs handling the solution space diversity have been proposed, such as GDEA \cite{ToffoloB03}, DEMO \cite{RobicF05}, DIVA \cite{UlrichBZ10}, ``MMEA'' \cite{ZhouZJ09}, DCMMMOEA \cite{XiaZY14}, and MOEA/D-EVSD \cite{CastilloSAML17}.
Note that solution space diversity management in these MOEAs aims to efficiently approximate the Pareto front for MOPs.
Since these methods were not designed for MMOPs, they are likely to perform poorly for MMOPs.
For example, ``MMEA'', which stands for a model-based multi-objective evolutionary algorithm, cannot find multiple equivalent Pareto optimal solutions  \cite{ZhouZJ09}.
Nevertheless,  helpful clues for designing an efficient MMEA can be found in these MOEAs.


The performance of MMEAs has not been well analyzed.
The post-processing method may perform better than MMEAs when the objective functions of a real-world problem are computationally expensive.
However, an in-depth investigation is necessary to determine which approach is more practical.
Whereas the population size $\mu$ and the maximum number of evaluations $n_{\rm max}$ were set to large values in some studies, they were set to small values in other studies.
For example, Table \ref{tab:emmoas} shows that $\mu = 1\,000$ and $n_{\rm max} = 500\,000$ for Omni-optimizer, while $\mu = 50$ and $n_{\rm max} = 50\,000$ for Niching-CMA.
It is unclear whether an MMEA designed with large $\mu$ and $n_{\rm max}$ values works well with small $\mu$ and $n_{\rm max}$ values. 
While MMOPs with four or more objectives appear in real-world applications (e.g., five-objective rocket engine design problems \cite{KudoYF11}), most MMEAs have been applied to only two-objective MMOPs.
A large-scale benchmarking study is necessary to address the above-mentioned issues. 

The decision maker may want to examine diverse dominated solutions.
As explained in Section \ref{sec:introduction}, dominated solutions found by $P_{Q, \epsilon}$-MOEA support the decision making in space mission design problems \cite{SchutzeVC11}.
The results presented in \cite{KrmicekS06} showed that diverse solutions found by 4D-Miner help neuroscientists analyze brain imaging data.
Although most MMEAs assume MMOPs with $\delta=0$ as shown in Table \ref{tab:emmoas}, MMEAs that can handle MMOPs with $\delta>0$ may be more practical.
Since most MMEAs (e.g., Omni-optimizer) remove dominated individuals from the population, they are unlikely to find diverse dominated solutions.
Some specific mechanisms are necessary to handle MMOPs with $\delta>0$ (e.g., the multiple subpopulation scheme in DIOP and MNCA).

As explained at the beginning of this section, MMEAs need the three abilities (1)--(3).
While the abilities (1) and (2) are needed to approximate the Pareto front, the ability (3) is needed to find equivalent Pareto optimal solutions.
Most existing studies (e.g., \cite{DebT08,ShirPNE09,YueQL17,TanabeI18}) report that the abilities (1) and (2) of MMEAs are worse than those of MOEAs.
For example, the results presented in \cite{TanabeI18} showed that Omni-optimizer, MO\_Ring\_PSO\_SCD, and MOEA/D-AD perform worse than NSGA-II in terms of IGD \cite{CoelloS04} (explained in Section \ref{sec:metrics}).
If the decision maker is not interested in the distribution of solutions in the solution space, it would be better to use MOEAs rather than MMEAs.
The poor performance of MMEAs for multi-objective optimization is mainly due to the ability (3), which prevents MMEAs from directly approximating the Pareto front.
This undesirable performance regarding the abilities (1) and (2) is an issue in MMEAs.

\noindent $\bullet$ {\it What to learn from MSOPs:}
An online data repository (\url{https://github.com/mikeagn/CEC2013}) that provides results of optimizers on the CEC2013 problem suite \cite{LiEE13} is available for MSOPs.
This repository makes the comparison of optimizers easy, facilitating constructive algorithm development.
A similar data repository is needed for studies of MMOPs.

The number of maintainable individuals in the population/archive strongly depends on the population/archive size.
However, it is usually impossible to know the number of equivalent Pareto optimal solutions of an MMOP a priori.
The same issue can be found in MSOPs.
To address this issue, the latest optimizers (e.g., dADE \cite{EpitropakisLB13} and RS-CMSA \cite{AhrariDP17}) have an unbounded archive that maintains solutions found during the search process.
Unlike modern optimizers for MSOPs, Table \ref{tab:emmoas} shows that only three MMEAs have such a mechanism.
The adaptive population sizing mechanisms in omni-aiNet, $P_{Q, \epsilon}$-MOEA, and MOEA/D-AD are advantageous.
A general strategy of using an unbounded (external) archive could improve the performance of MMEAs.





\section{Multi-modal multi-objective test problems}
\label{sec:test_problems}

\begin{figure}[t]
\newcommand{\widthvar}{0.17}
\centering
\subfloat[SYM-PART1]{\includegraphics[width=0.161\textwidth]{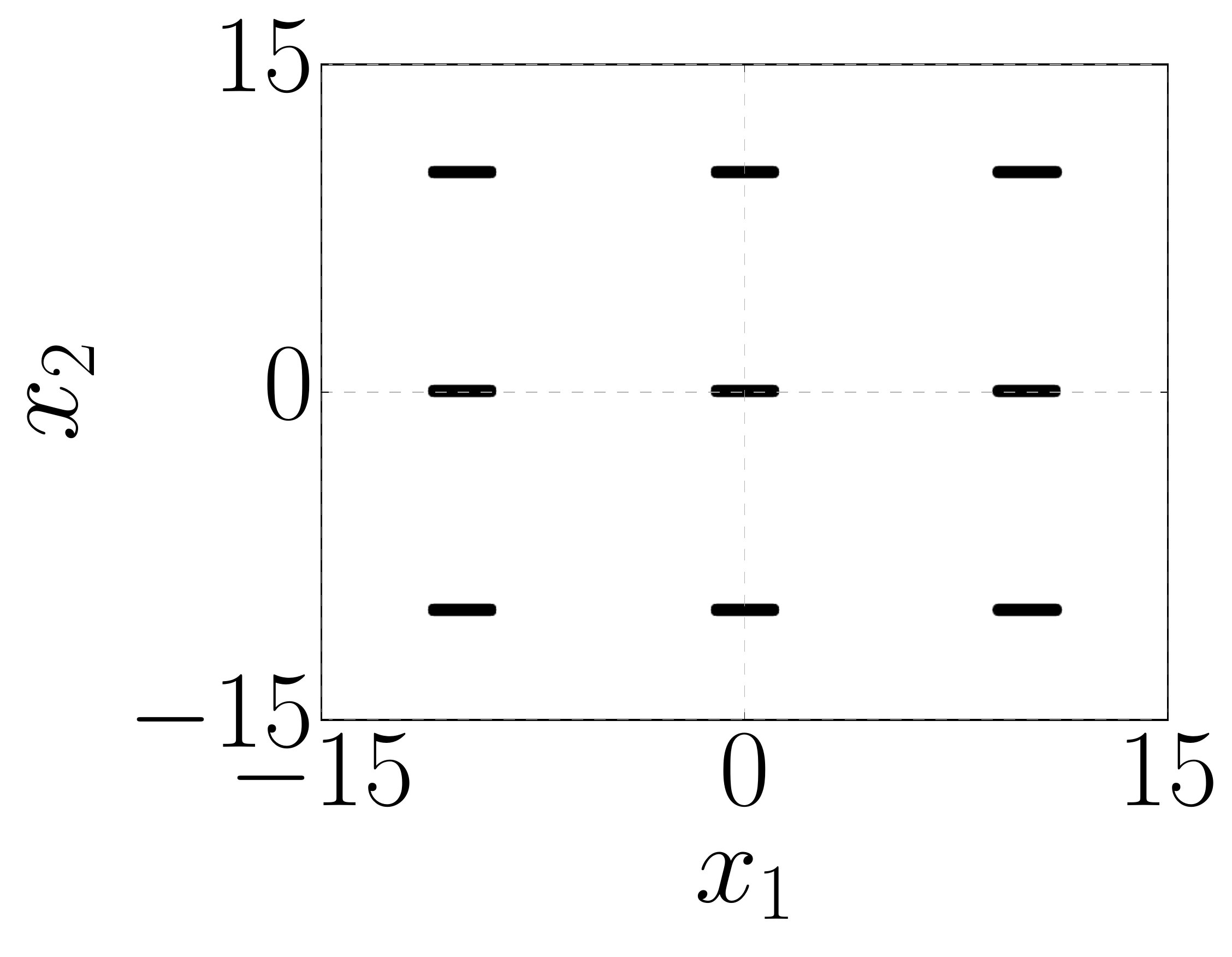}}
\subfloat[SYM-PART2]{\includegraphics[width=0.163\textwidth]{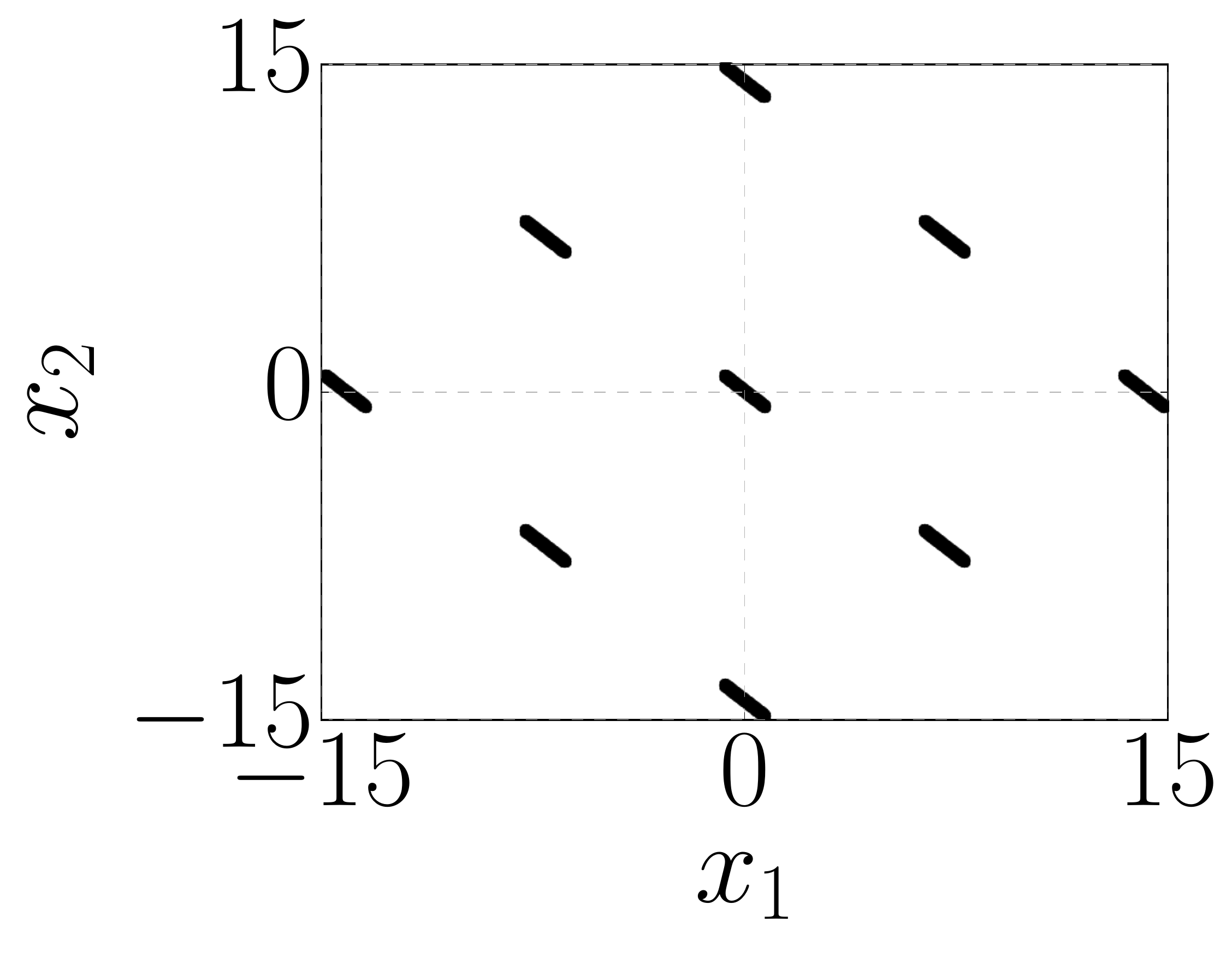}}
\subfloat[SYM-PART3]{\includegraphics[width=0.161\textwidth]{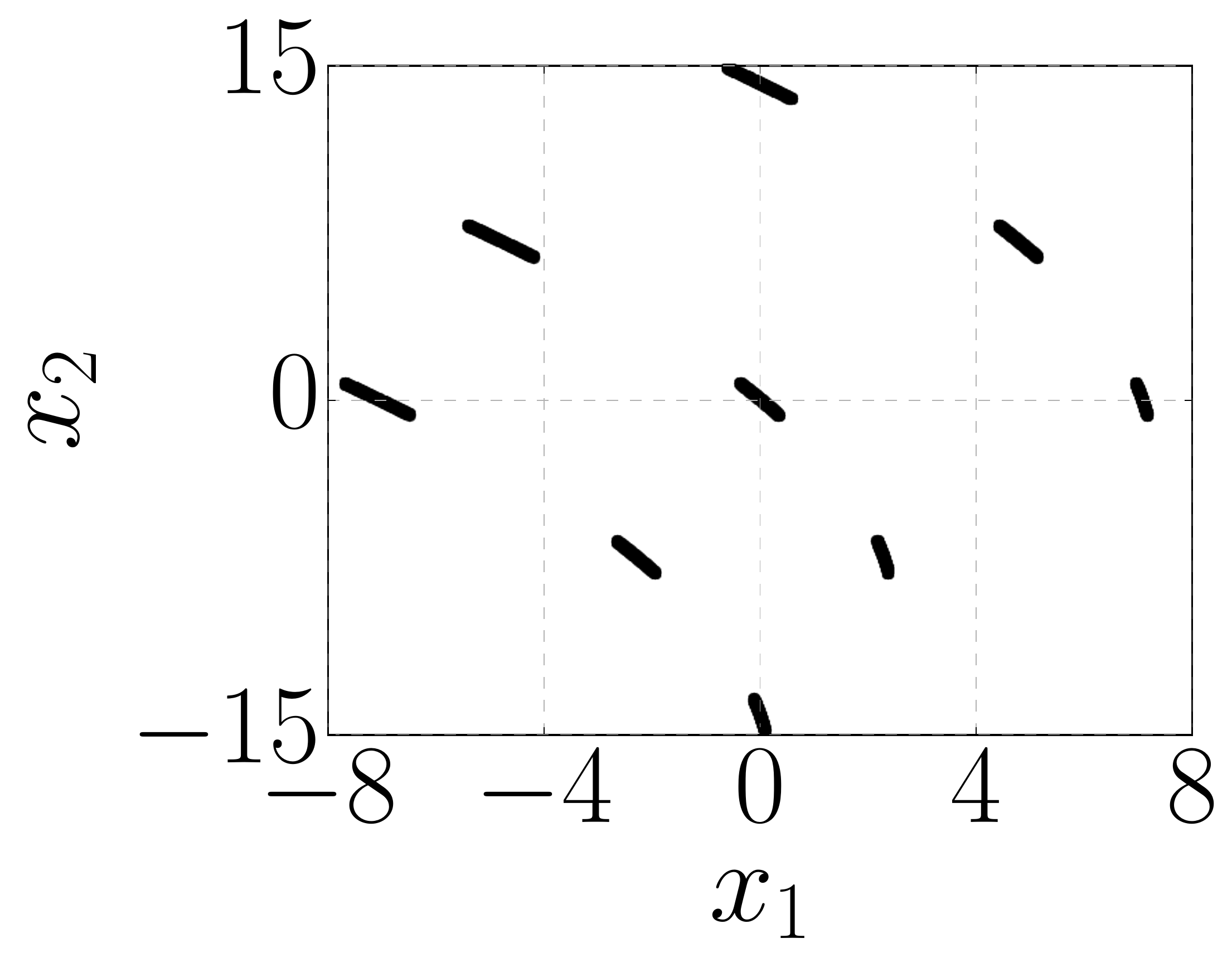}}
\\[-2mm]
\subfloat[Two-On-One]{\includegraphics[width=0.17\textwidth]{graph/ps_2d_distribution_var/TWO-ON-ONE.pdf}}
\subfloat[Omni-test]{\includegraphics[width=0.159\textwidth]{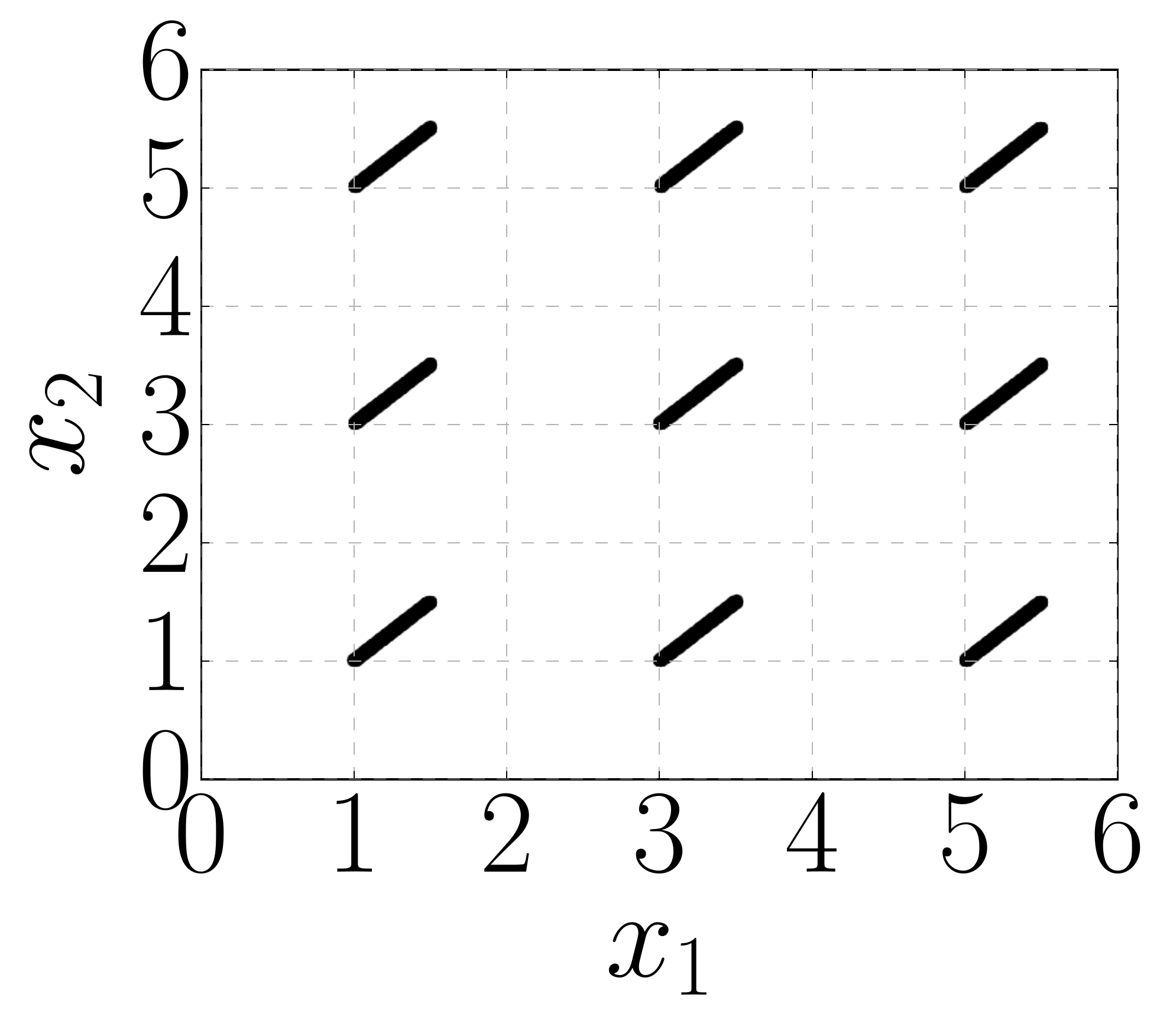}}
\subfloat[Superspheres]{\includegraphics[width=0.166\textwidth]{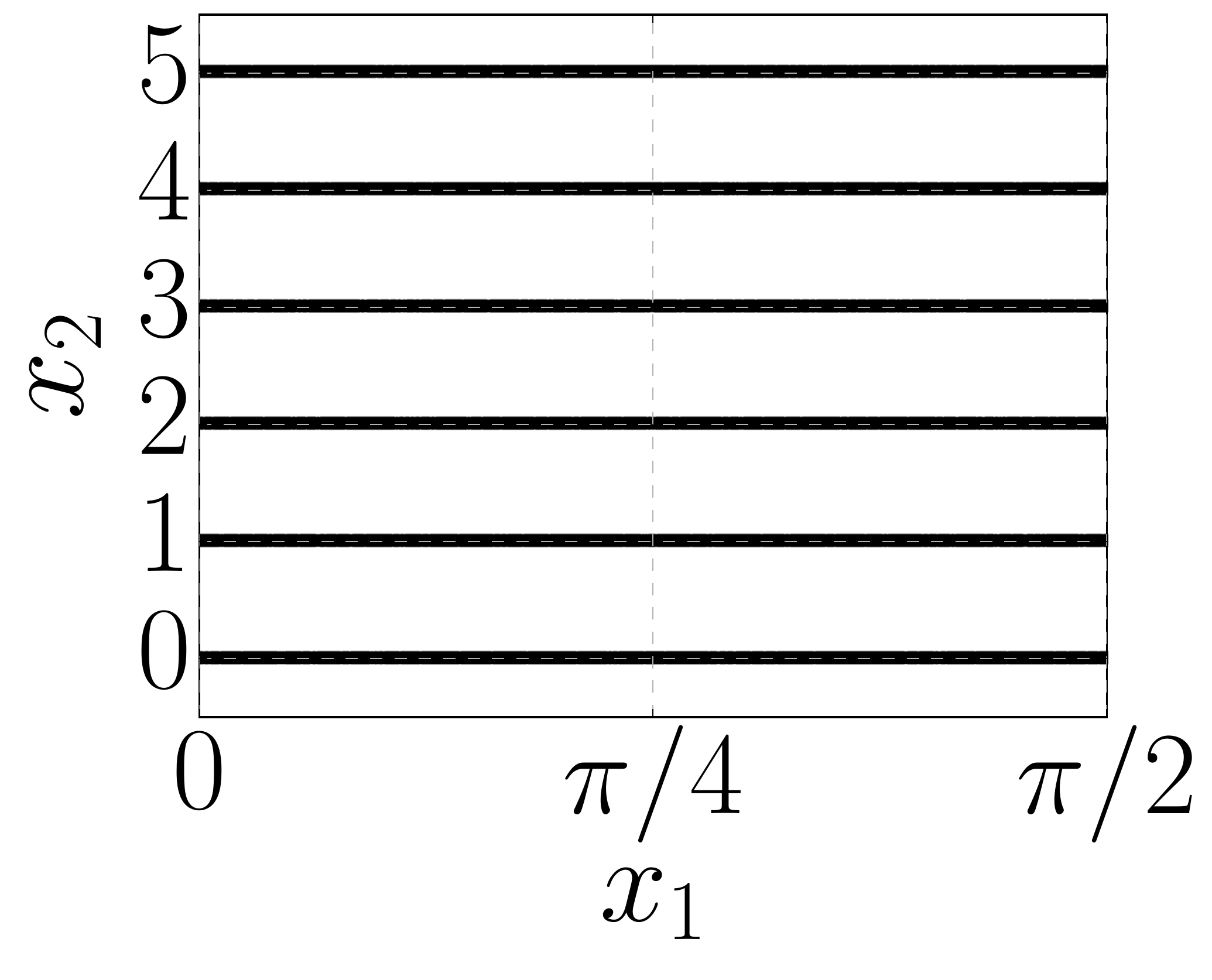}}
\\[-2mm]
\subfloat[SSUF1]{\includegraphics[width=0.161\textwidth]{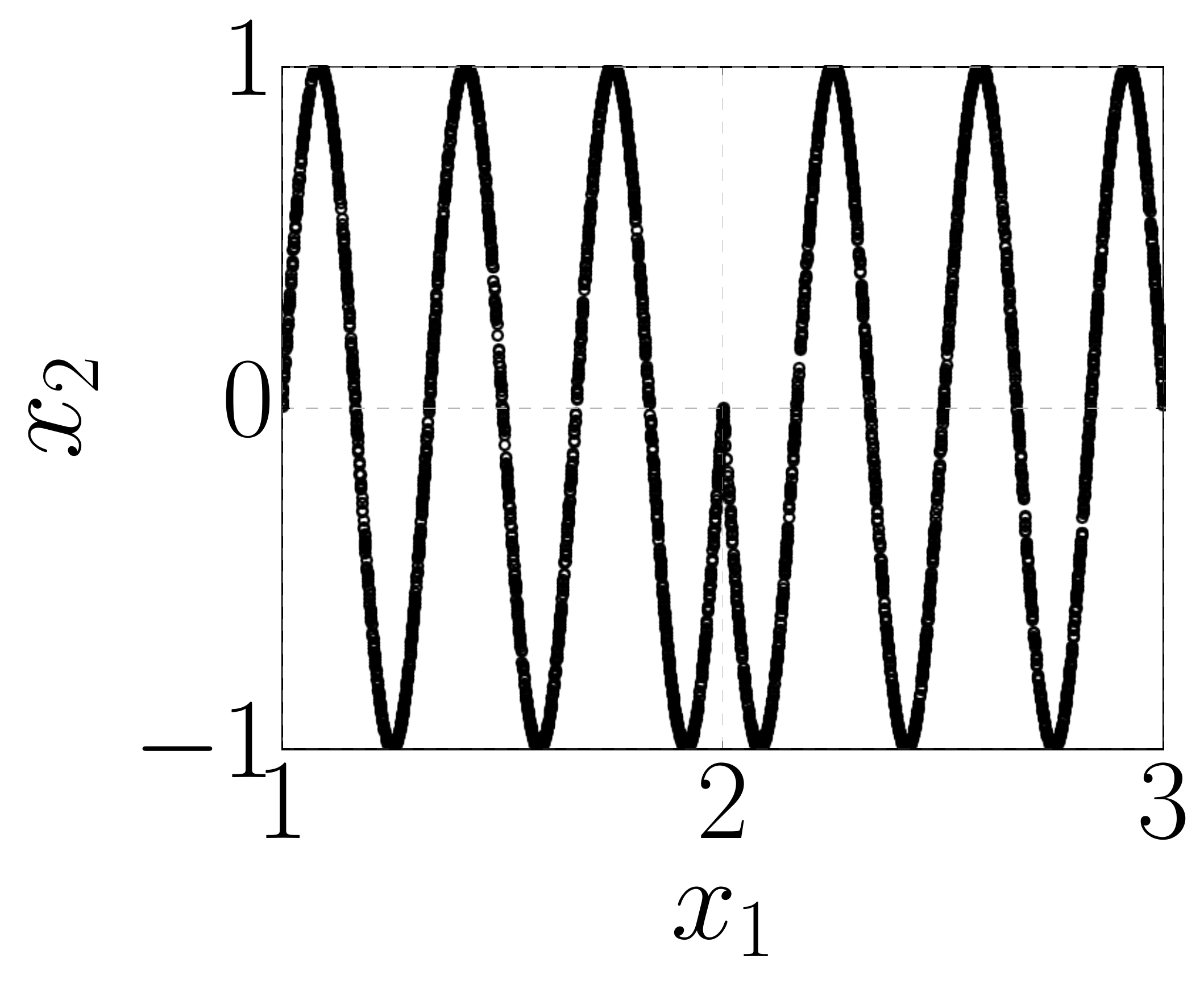}}
\subfloat[SSUF3]{\includegraphics[width=0.16\textwidth]{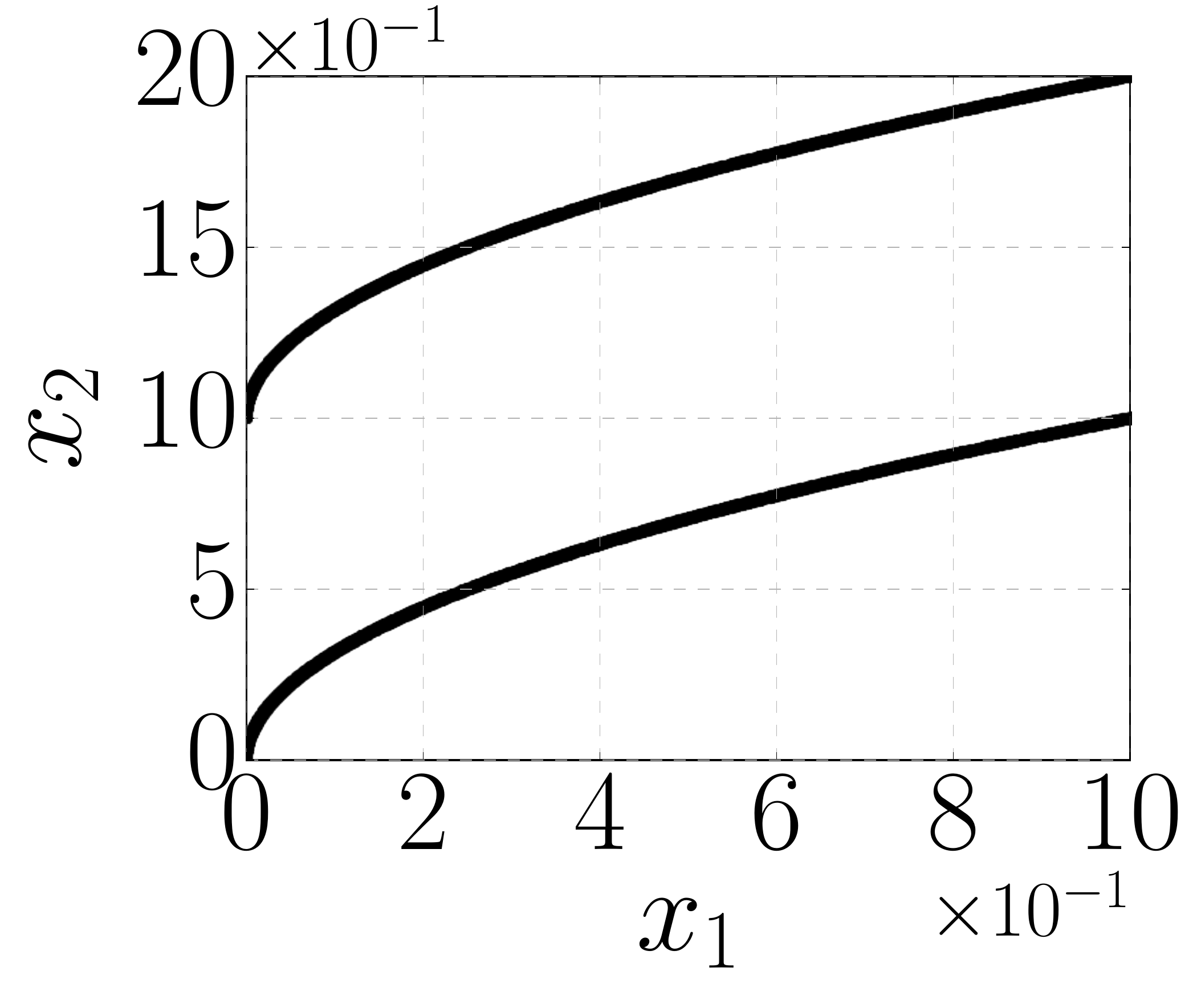}}
\subfloat[Polygon]{\includegraphics[width=0.162\textwidth]{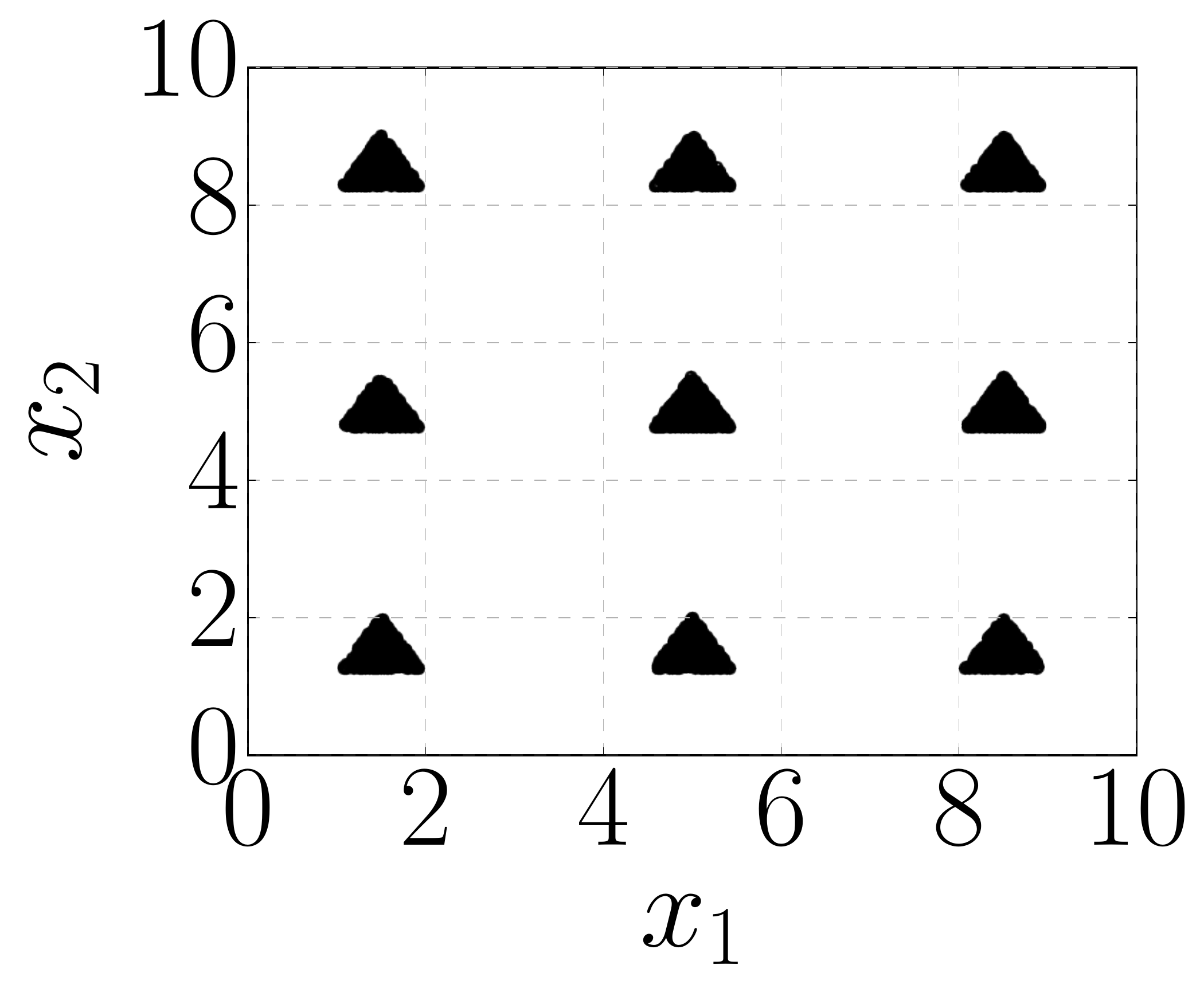}}
\caption{
\small
Distribution of the Pareto optimal solutions for the eight problems.
Only $x_1$ and $x_2$ are shown on Omni-test.
}
\label{fig:mmop_test_problems}
\end{figure}

This section describes test problems for benchmarking MMEAs.
Unlike multi-objective test problems (e.g., the DTLZ \cite{DebTLZ05} test suite), multi-modal multi-objective test problems were explicitly designed such that they have multiple equivalent Pareto optimal solution subsets.
The two-objective and two-variable SYM-PART1 \cite{RudolphNP07} is one of the most representative test problems for benchmarking MMEAs: $f_1(\vector{y}) = (y_1 + a)^2 + y_2^2$ and $f_2(\vector{y}) = (y_1 - a)^2 + y_2^2$.
Here, $y_1$ and $y_2$ are translated values of $x_1$ and $x_2$ as follows: $y_1 = x_1 - t_1 (c+2a)$ and $y_2 = x_2 -t_2 b$.
In SYM-PART1, $a$ controls the region of Pareto optimal solutions, and $b$ and $c$ specify the positions of the Pareto optimal solution subsets.
The so-called tile identifiers $t_1$ and $t_2$ are randomly selected from $\{-1, 0, 1\}$.
Fig. \ref{fig:mmop_test_problems}(a) shows the shape of the Pareto optimal solutions of SYM-PART1 with $a=1$, $b=10$, and $c=8$.
As shown in Fig. \ref{fig:mmop_test_problems}(a), the equivalent Pareto optimal solution subsets are on nine lines in SYM-PART1.

Other test problems include the Two-On-One \cite{PreussNR06} problem, the Omni-test problem \cite{DebT08}, the SYM-PART2 and SYM-PART3 problems \cite{RudolphNP07}, the Superspheres problem \cite{EmmerichD06}, the EBN problem \cite{BeumeNE07}, the two SSUF problems \cite{LiangYQ16}, and the Polygon problems \cite{IshibuchiHTN10}. 
Fig. \ref{fig:mmop_test_problems} also shows the distribution of their Pareto optimal solutions.
Since there are an infinite number of Pareto optimal solutions in the EBN problem, we do not show them.
Source codes of the ten problems can be downloaded from the supplementary website (\url{https://sites.google.com/view/emmo/}).
%
%
In Omni-test, equivalent Pareto optimal solution subsets are regularly located.
%
SYM-PART2 is a rotated version of SYM-PART1.
SYM-PART3 is a transformed version of SYM-PART2 using a distortion operation.
%
The Superspheres problem with $D=2$ has six equivalent Pareto optimal solution subsets.
However, the number of its $P$ is unknown for $D>2$.
EBN can be considered as a real-coded version of the so-called binary one-zero max problem.
All solutions in the solution space are Pareto optimal solutions.
%
SSUF1 and SSUF3 are extensions of the UF problems \cite{ZhangZZSLT08} to MMOPs.
There are two symmetrical Pareto optimal solution subsets in SSUF1 and SSUF3.
Polygon is an extension of the distance minimization problems \cite{KoppenY07} to MMOPs, where $P$ equivalent Pareto optimal solution subsets are inside of $P$ regular $M$-sided polygons.

In addition, the eight MMF problems are presented in \cite{YueQL17}.
Similar to SSUF1 and SSUF3, the MMF problems are derived from the idea of designing a problem that has multiple equivalent Pareto optimal solution subsets by mirroring the original one.
A bottom-up framework for generating scalable test problems with any $D$ is proposed in  \cite{ZhangSA17}.
$P$ equivalent Pareto optimal solution subsets are in $P$ hyper-rectangular located in the solution space similar to the SYM-PART problems.
While the first $k$ variables play the role of ``position'' parameters in the solution space, the other $D-k$ variables represent ``distance'' parameters.
The six HPS problem instances were constructed using this framework in \cite{ZhangSA17}.


If a given problem has the multi-modal fitness landscape, it may have multiple non-Pareto fronts whose shapes are similar to the true Pareto front.
Such a problem (e.g., ZDT4 \cite{ZitzlerDT00}) is referred to as a multi-frontal test problem \cite{HubandHBW06}.
If the $\delta$ value (defined in Subsection \ref{sec:def_mmops}) is sufficiently large, a multi-frontal test problem can be regarded as a multi-modal multi-objective test problem.
In fact, ZDT4 was used in \cite{ZechmanGS13} as a test problem.
%
The Kursawe problem \cite{Kursawe90} is a multi-modal and nonseparable test problem with a disconnected Pareto front.
The Kursawe problem has two fronts in the objective space similar to multi-frontal problems.
Thus, the Kursawe problem can be used as a multi-modal multi-objective test problem.

\begin{table}[t]
\begin{center}
  \caption{\small Properties of multi-modal multi-objective test problems, where $M$, $D$, and $P$ denote the number of objectives, design variables, and equivalent Pareto optimal solution subsets, respectively.
If a problem has irregularity, the shapes of its multiple equivalent Pareto optimal solution subsets differ from each other.
}
{\footnotesize
  \label{suptab:mmop_test_problems}
\scalebox{1}[1]{ 
\begin{tabular}{lccccc}
\midrule
Test problems & $M$ & $D$ & $P$ & Irregularity\\ 
\toprule
SYM-PART problems \cite{RudolphNP07} & 2  & 2 & 9 & $\checkmark$\\\midrule
Two-On-One problem \cite{PreussNR06} & 2  & 2 & 2 & \\\midrule
Omni-test problem \cite{DebT08} & 2 & Any & $3^D$ & \\\midrule
Superspheres problem \cite{EmmerichD06} & 2 & Any & Unknown & \\\midrule
EBN problem \cite{BeumeNE07} & 2  & Any & $\infty$ & \\\midrule
Polygon problems \cite{IshibuchiHTN10} & Any  & 2 & Any & \\\midrule
SSUF problems \cite{LiangYQ16} & 2  & 2 & 2 & \\\midrule
MMF suite \cite{YueQL17} & 2  & 2 & 2 or 4 & \\\midrule
HPS suite  \cite{ZhangSA17} & 2  & Any & Any & \\\midrule
\end{tabular}
}
}
\end{center}
\end{table}

\subsubsection{Open issues}
\label{section:problems_open_issues}


Table \ref{suptab:mmop_test_problems} summarizes the properties of multi-modal multi-objective test problems reviewed here.
In Table \ref{suptab:mmop_test_problems}, $P$ of Omni-test adheres to \cite{UlrichBT10}.

%
Table \ref{suptab:mmop_test_problems} indicates that scalable test problems do not exist, in terms of $M$, $D$, and $P$.
Although the SYM-PART problems have some desirable properties (e.g., their adjustable and straightforward Pareto optimal solution shapes), $M$, $D$, and $P$ are constant in these problems.
%
%
Only Polygon is scalable in $M$. 
While most test problems have only two design variables, Omni-test and HPS are scalable in $D$.
Unfortunately, $P$ increases exponentially with increased $D$ in Omni-test due to the combinatorial nature of variables.
Although the idea of designing scalable SYM-PART and Polygon problems to $D$ is presented in \cite{HuangQDZSLPH07,IshibuchiYAN13}, they have similar issues to Omni-test.
Although the HPS problems do not have such an issue, it is questionable whether there exists a real-world problem with design variables affecting only the distance between the objective vectors and the Pareto front.
%
%
Only SYM-PART3 has irregularity.
Since the shapes of the Pareto optimal solution subsets may be different from each other in real-world problems, we believe that test problems with the irregularity are necessary to evaluate the performance of MMEAs.
The performance of an MMEA with an absolutely defined niching radius (e.g., DNEA) is likely to be overestimated in test problems without irregularity.

%

In addition, the relation between synthetic test problems and real-world problems has not been discussed.
The idea of designing a Polygon problem based on a real-world map is presented in \cite{IshibuchiAN11}.
However, this does not mean that such a Polygon problem is an actual real-world problem.

\noindent $\bullet$ {\it What to learn from MSOPs:}
Some construction methods for multi-modal single-objective test problems are available, such as the software framework proposed in \cite{RonkkonenLKL11}, the construction method for various problems \cite{QuLWCS16}, and Ahrari and Deb's method \cite{AhrariD18}.
Borrowing ideas from such sophisticated construction methods is a promising way to address the above-mentioned issues of multi-modal multi-objective test problems.
In \cite{RonkkonenLKL11}, R{\"{o}}nkk{\"{o}}nen et al. present eight desirable properties for multi-modal single-objective problem generators such as scalability in $D$, control of the number of global and local optima, and regular and irregular distributions of optima.
These eight properties can be a useful guideline for designing multi-modal multi-objective problem generators.





\section{Performance indicators for MMEAs} 
\label{sec:metrics}

Performance indicators play an important role in quantitatively evaluating the performance of MOEAs as well as MMEAs.
Since performance indicators for MOEAs consider only the distribution of objective vectors (e.g., the hypervolume, GD, and IGD indicators \cite{ZitzlerTLFF03,CoelloS04}), they cannot be used to assess the ability of MMEAs to find multiple equivalent Pareto optimal solutions.
For this reason, some indicators have been specially designed for MMEAs.
Performance indicators for MMEAs can be classified into two categories: simple extensions of existing performance indicators for MOEAs and specific indicators based on the distributions of solutions.


IGDX \cite{SchutzeVC11,ZhouZJ09} is a representative example of the first approach.
The IGD and IGDX indicators are given as follows:
{\small
\begin{align}
 \label{eqn:igd}
       {\rm IGD} (\vector{A}) &= \frac{1}{|\vector{A}^*|} \left(\sum_{\vector{z} \in \vector{A}^*} \min_{\vector{x} \in \vector{A}} \Bigl\{ {\rm ED} \bigl(\vector{f}(\vector{x}), \vector{f}(\vector{z}) \bigr) \Bigr\} \right), 
\\  
\label{eqn:igdx}
{\rm IGDX} (\vector{A}) &= \frac{1}{|\vector{A}^*|} \left(\sum_{\vector{z} \in \vector{A}^*} \min_{\vector{x} \in \vector{A}} \Bigl\{ {\rm ED} \bigl(\vector{x}, \vector{z} \bigr) \Bigr\} \right), 
 \end{align}
}%
where $\vector{A}$ is a set of solutions obtained by an MMEA and $\vector{A}^*$ is a set of reference solutions in the Pareto optimal solution set.
${\rm ED}(\vector{x}_1, \vector{x}_2)$ denotes the Euclidean distance between $\vector{x}_1$ and $\vector{x}_2$.
%
While $\vector{A}$ with a small IGD value is a good approximation of the Pareto front, $\vector{A}$ with a small IGDX approximates Pareto optimal solutions well.
Other indicators in the first category include GDX \cite{SchutzeVC11}, the Hausdorff distance indicator \cite{SchutzeELC12} in the solution space \cite{SchutzeVC11},  CR \cite{YueQL17}, and PSP \cite{YueQL17}.
GDX is a GD indicator in the solution space similar to IGDX.
CR is an alternative version of the maximum spread \cite{ZitzlerTLFF03} to measure the spread of $\vector{A}$.
PSP is a combination of IGDX and CR. 

Performance indicators in the second category include the mean of the pairwise distance between two solutions \cite{ShirPNE09}, CS \cite{RudolphNP07}, SPS \cite{RudolphNP07}, the Solow-Polasky diversity measure \cite{SolowP94} used in \cite{IshibuchiYAN12,UlrichBT10}, and PSV \cite{ZhangSA17}.
CS is the number of Pareto optimal solution subsets covered by at least one individual.
SPS is the standard deviation of the number of solutions close to each Pareto optimal solution subset.
PSV is the percentage of the volume of $\vector{A}$ in the volume of $\vector{A}^*$ in the solution space.






\begin{table}[t]
\renewcommand{\arraystretch}{0.8}
\begin{center}
  \caption{Properties of performance indicators for MMEAs (convergence to Pareto optimal solution subsets, diversity, uniformity, spread, the use of reference solution sets, and possibility to compare solution sets with different sizes).}
{\footnotesize
  \label{tab:performance_indicators}
\scalebox{1}[1]{ 
\begin{tabular}{lcccccccc}
\midrule
Indicators & Conv. & Div. & Unif. & Spr. & Ref. & Dif.\\
\toprule
GDX \cite{SchutzeVC11} & $\checkmark$ &  &  & &  $\checkmark$ & \\\midrule
IGDX \cite{SchutzeVC11,ZhouZJ09} & $\checkmark$ & $\checkmark$ & $\checkmark$ & $\checkmark$ & $\checkmark$ & \\\midrule
Hausdorff distance \cite{SchutzeVC11}& $\checkmark$ & $\checkmark$ &  & $\checkmark$ & $\checkmark$ & $\checkmark$\\\midrule
CR \cite{YueQL17}&  & $\checkmark$ & &  $\checkmark$ & $\checkmark$ & \\\midrule
PSP \cite{YueQL17} & $\checkmark$ & $\checkmark$ & $\checkmark$ & $\checkmark$ & $\checkmark$ & \\\midrule
Pairwise distance \cite{ShirPNE09} & & $\checkmark$ &  & $\checkmark$ &  & \\\midrule
CS \cite{RudolphNP07} & $\checkmark$ & $\checkmark$ & $\checkmark$ & $\checkmark$ & $\checkmark$ & \\\midrule
SPS \cite{RudolphNP07} & $\checkmark$ & $\checkmark$ & $\checkmark$ &  & $\checkmark$ & \\\midrule
Solow-Polasky \cite{SolowP94} &  & $\checkmark$ & $\checkmark$ & $\checkmark$  & & $\checkmark$\\\midrule
PSV \cite{ZhangSA17} & $\checkmark$ & $\checkmark$ &  & $\checkmark$ & $\checkmark$ & $\checkmark$\\\midrule
\end{tabular}
}
}
 \end{center}
\end{table}

\begin{figure}[t]
\newcommand{\widthvar}{0.24}
  \begin{center} 
    \subfloat[$\vector{A}_1$ in the solution space]{\includegraphics[width=0.23\textwidth]{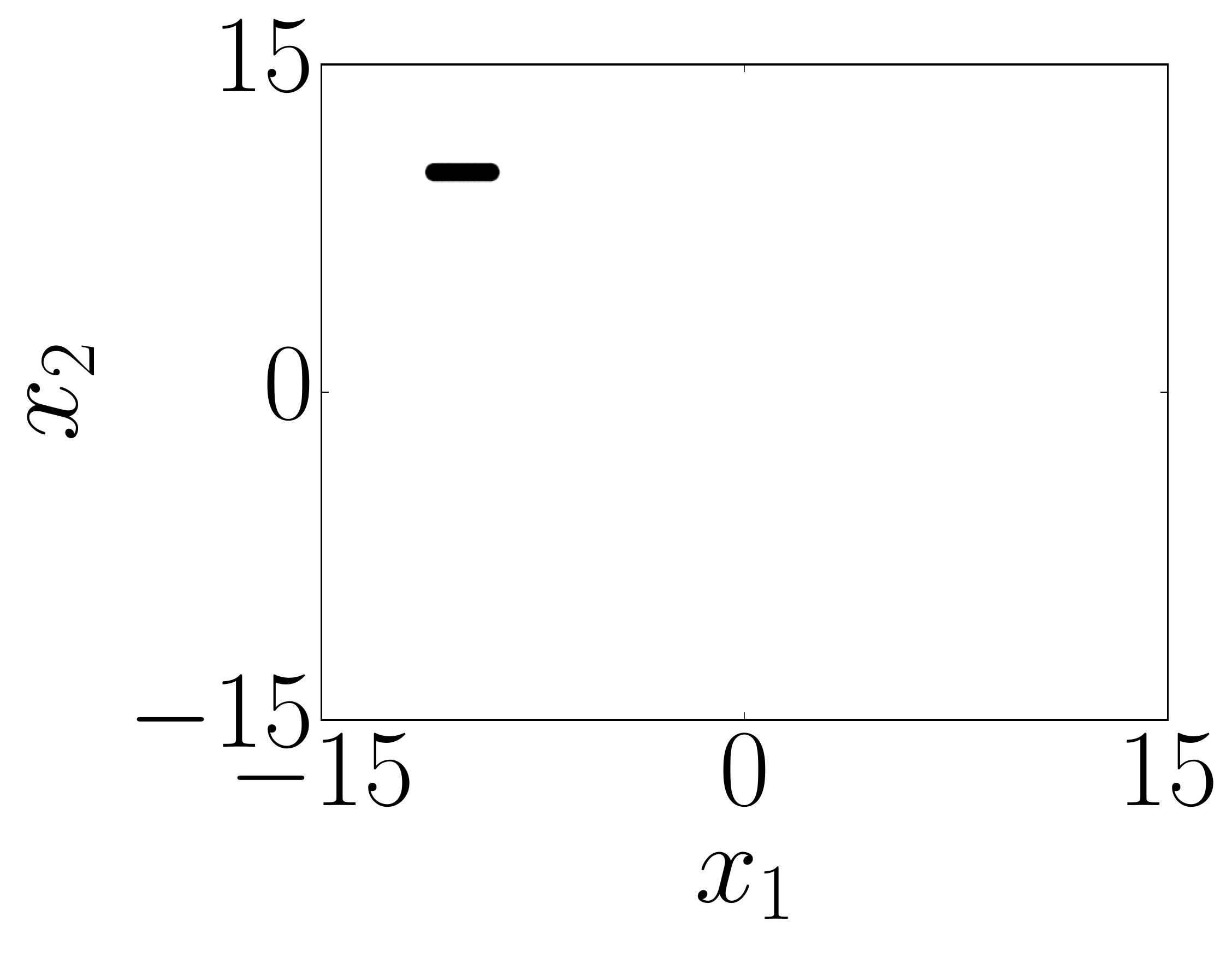}}
    \subfloat[$\vector{A}_2$ in the solution space]{\includegraphics[width=0.23\textwidth]{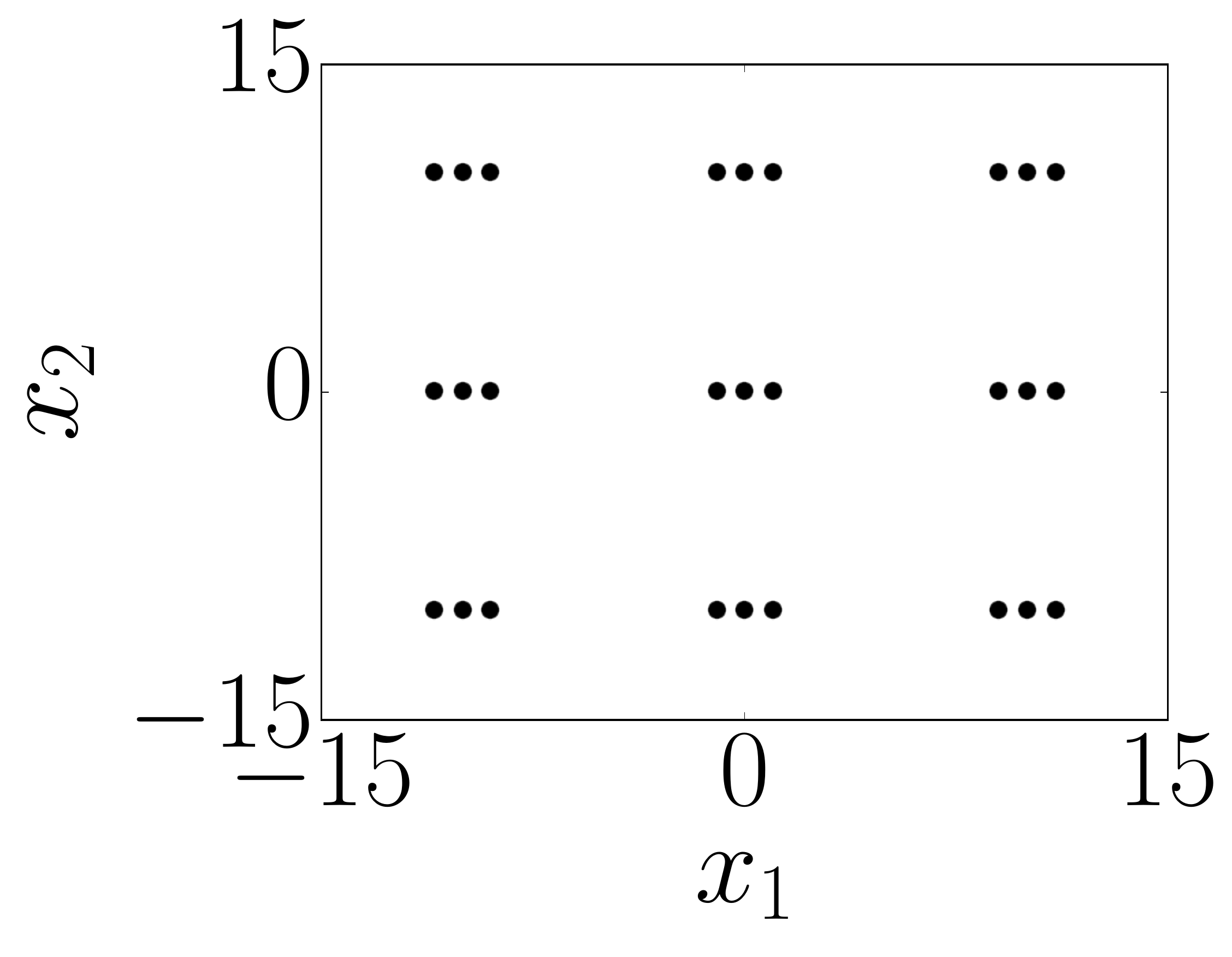}}\\
    \subfloat[$\vector{A}_1$ in the objective space]{\includegraphics[width=0.2\textwidth]{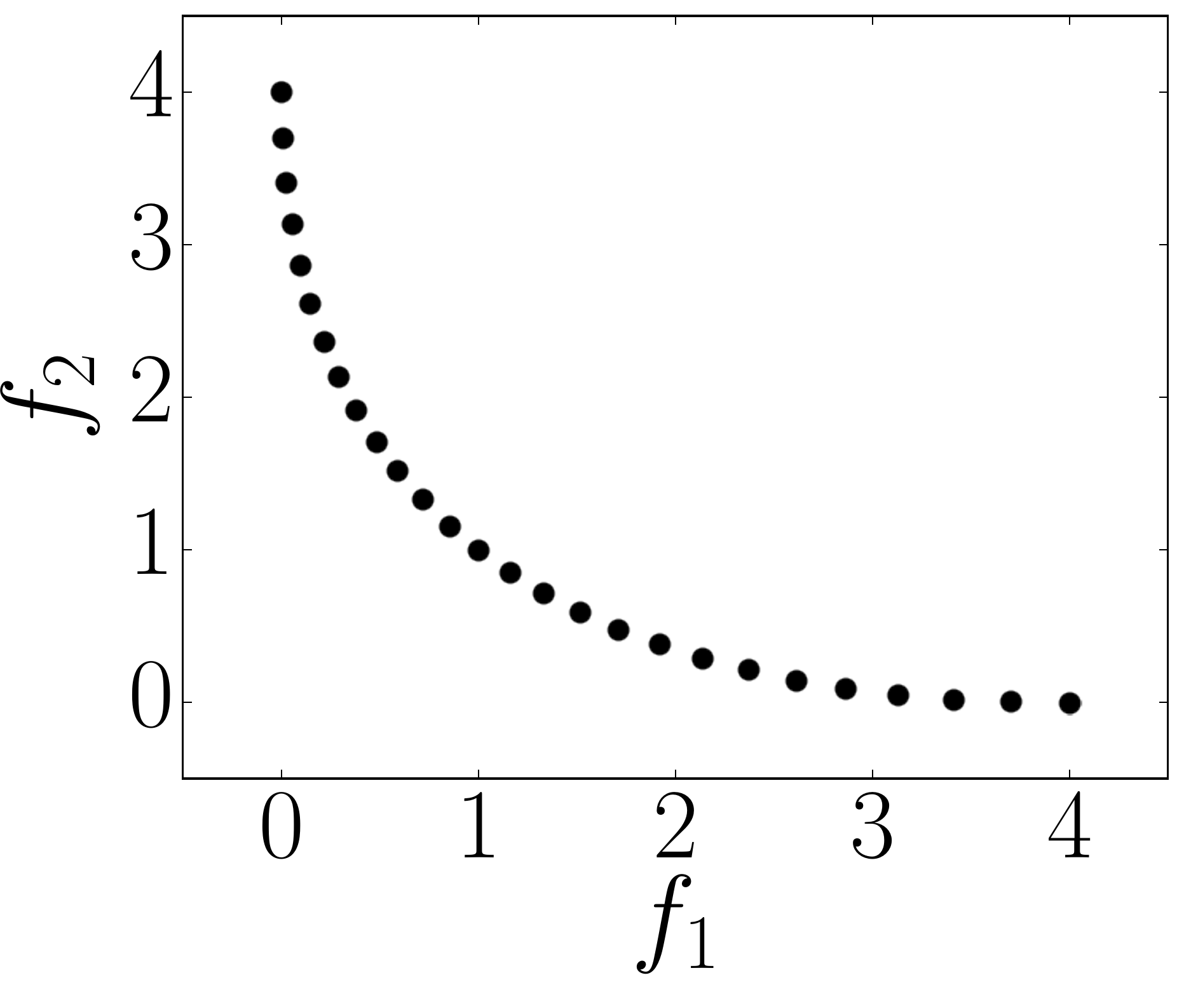}}
    \hspace{2mm}
    \subfloat[$\vector{A}_2$ in the objective space]{\includegraphics[width=0.201\textwidth]{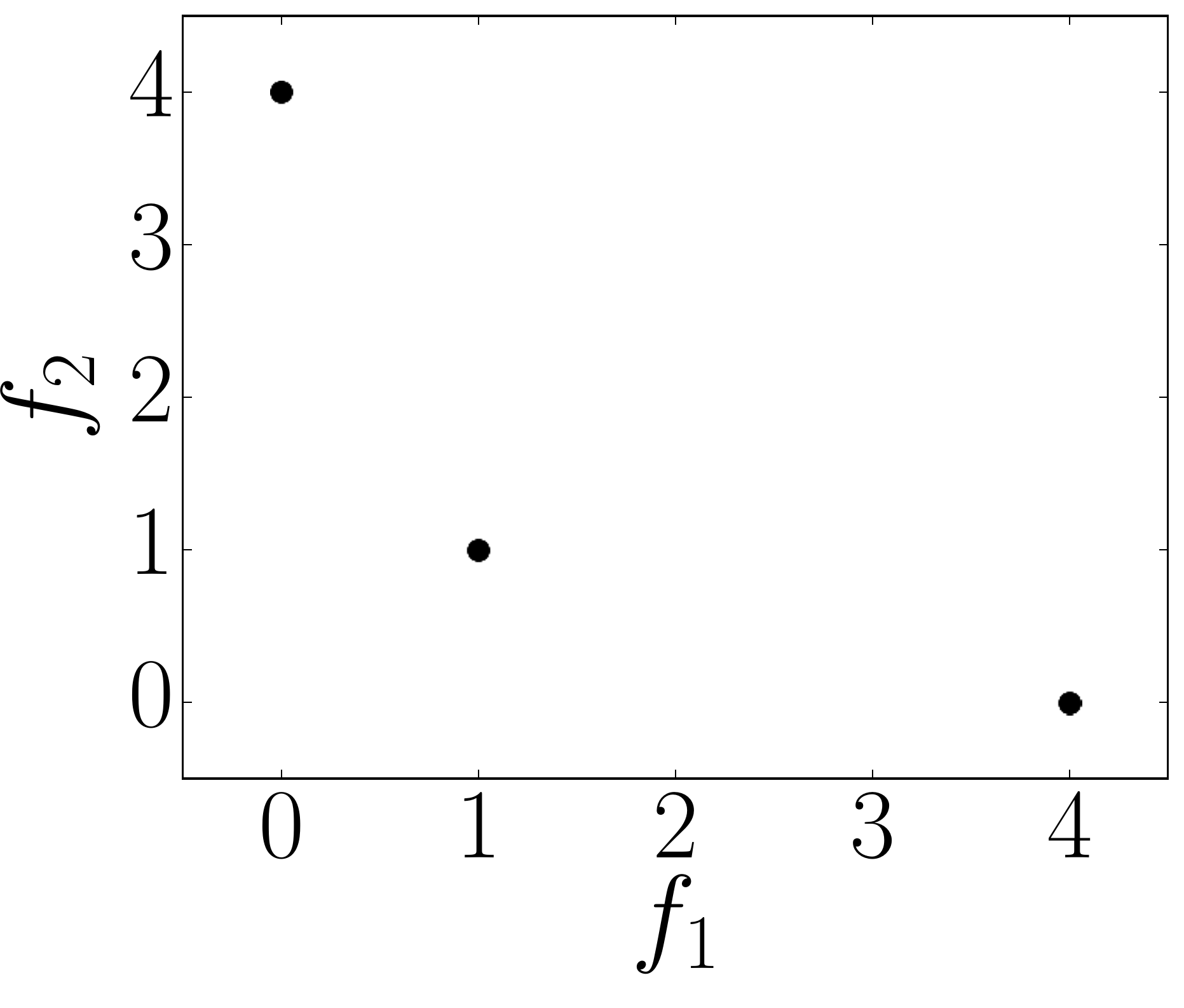}}
    \caption{
\small
  Comparison of solution sets $\vector{A}_1$ and $\vector{A}_2$  for SYM-PART1.  
}
\label{fig:igd_igdx}
  \end{center}
\end{figure}

\subsubsection{Open issues}

%


Table \ref{tab:performance_indicators} shows the properties of performance indicators for MMEAs reviewed in this section, where the properties are assessed based on the description of each indicator.
While the properties of the performance indicators for MOEAs have been examined (e.g., \cite{ZitzlerTLFF03,SchutzeELC12}), those for MMEAs have not been well analyzed.

Performance indicators for MMEAs should be able to evaluate the three abilities (1)--(3) explained in Section \ref{sec:emmo}.
Although IGDX is frequently used, it should be noted that IGDX does not evaluate the distribution of solutions in the objective space.
Fig. \ref{fig:igd_igdx} shows the distribution of two solution sets $\vector{A}_1$ and $\vector{A}_2$ for SYM-PART1 in the solution and objective spaces, where $|\vector{A}_1|$ and $|\vector{A}_2|$ are 27.
While the solutions in $\vector{A}_1$ are evenly distributed on one of the nine Pareto optimal solution subsets, the solutions in $\vector{A}_2$ are evenly distributed on all of them.
Although $\vector{A}_1$ has 27 objective vectors that cover the Pareto front, $\vector{A}_2$ has only 3 equivalent objective vectors.
The IGDX and IGD values of $\vector{A}_1$ and $\vector{A}_2$ are as follows: ${\rm IGDX}(\vector{A}_1) = 15.92$, ${\rm IGDX}(\vector{A}_2) = 0.25$, ${\rm IGD}(\vector{A}_1) = 0.06$, and ${\rm IGD}(\vector{A}_2) = 0.81$.
We used $5\,000$ Pareto optimal solutions for $\vector{A}^*$.
Although $\vector{A}_2$ has a worse distribution in the objective space than $\vector{A}_1$, ${\rm IGDX}(\vector{A}_2)$ is significantly better than ${\rm IGDX}(\vector{A}_1)$.
As demonstrated here, IGDX can evaluate the abilities (1) and (3) but cannot evaluate the ability (2) to find diverse solutions in the objective space.
%
Since the other indicators in Table \ref{tab:performance_indicators} do not take into account the distribution of objective vectors similar to IGDX, they are likely to have the same undesirable property.
For a fair performance comparison, it is desirable to use the indicators for MOEAs (e.g., hypervolume and IGD) in addition to the indicators for MMEAs in Table \ref{tab:performance_indicators}.

\noindent $\bullet$ {\it What to learn from MSOPs:}
It is desirable that the indicators for multi-modal single-objective optimizers evaluate a solution set without the knowledge of the fitness landscape such as the positions of the optima and the objective values of the optima \cite{MwauraEN16}.
The same is true for indicators for MMEAs.
Table \ref{tab:performance_indicators} shows that most indicators (e.g., IGDX) require $\vector{A}^*$.
Since $\vector{A}^*$ is usually unavailable in real-world problems,
it is desirable that indicators for MMEAs evaluate $\vector{A}$ without $\vector{A}^*$.

Since the archive size in modern multi-modal single-objective optimizers is unbounded in order to store a number of local optima \cite{LiEDE17}, most indicators in this field can handle solution sets with different sizes (e.g., the peak ratio and the success rate \cite{LiEE13}).
For the same reason, it is desirable that indicators for MMEAs evaluate solution sets with different sizes in a fair manner.
However, it is difficult to directly use indicators for multi-modal single-objective optimizers to evaluate MMEAs.

\section{Conclusion}
\label{sec:conclusion}


The contributions of this paper are threefold.
The first contribution is that we reviewed studies in this field in terms of definitions of MMOPs, MMEAs, test problems, and performance indicators.
It was difficult to survey the existing studies of MMOPs for the reasons described in Section \ref{sec:introduction}.
Our review helps to elucidate the current progress on evolutionary multi-modal multi-objective optimization.
The second contribution is that we clarified open issues in this field.
In contrast to multi-modal single-objective optimization, multi-modal multi-objective optimization has not received much attention despite its practical importance.
Thus, some critical issues remain.
The third contribution is that we pointed out an issue associated with performance indicators for MMEAs.
Reliable performance indicators are necessary for the advancement of MMEAs.
We hope that this paper will encourage researchers to work in this research area, which is not well explored.

\section*{Acknowledgment}

This work was supported by the Program for Guangdong Introducing Innovative and Enterpreneurial Teams (Grant No. 2017ZT07X386), Shenzhen Peacock Plan (Grant No. KQTD2016112514355531), the Science and Technology Innovation Committee Foundation of Shenzhen (Grant No. ZDSYS201703031748284), the Program for University Key Laboratory of Guangdong Province (Grant No. 2017KSYS008), and National Natural Science Foundation of China (Grant No. 61876075).







\ifCLASSOPTIONcaptionsoff
  \newpage
\fi



%



\bibliography{reference}
\bibliographystyle{IEEEtran}

%








\end{document}